\documentclass{article}

\usepackage{arXiv/style_arxiv}

\usepackage[utf8]{inputenc} 
\usepackage[T1]{fontenc}    
\usepackage{hyperref}       
\usepackage{url}            
\usepackage{booktabs}       
\usepackage{amsfonts}       
\usepackage{nicefrac}       
\usepackage{microtype}      
\usepackage{xcolor}         

\usepackage{graphicx}
\usepackage{subcaption}
\DeclareCaptionSubType*[arabic]{figure}
\usepackage{amsmath}
\usepackage{amssymb}
\usepackage{mathtools}
\usepackage{amsthm}

\usepackage[ruled, vlined]{algorithm2e}
\SetKwInput{KwInput}{Input}
\SetKwInput{KwOutput}{Output}
\usepackage{algpseudocode}
\usepackage{algorithm2e}

\title{Mix from Failure: \\Confusion-Pairing Mixup for Long-Tailed Recognition}

\author{%
  Youngseok Yoon \\
  UCSB \\
  California, USA \\
  \texttt{youngseok\_yoon@ucsb.edu} \\
  \And
  Sangwoo Hong \\
  Seoul National University \\
  South Korea \\
  \texttt{swhong@cml.snu.ac.kr} \\
  \And
  Hyungjun Joo \\
  Seoul National University \\
  South Korea \\
  \texttt{joohj911@cml.snu.ac.kr} \\
  \AND
  Yao Qin \\
  UCSB, Google \\
  California, USA \\
  \texttt{yaoqin@ucsb.edu} \\
  \And
  Haewon Jeong \\
  UCSB \\
  California, USA \\
  \texttt{haewon@ucsb.edu} \\
  \And
  Jungwoo Lee \\
  Seoul National University \\
  South Korea \\
  \texttt{junglee@snu.ac.kr}
}

\begin{document}

\maketitle

\begin{abstract}
Long-tailed image recognition is a computer vision problem considering a real-world class distribution rather than an artificial uniform. Existing methods typically detour the problem by i) adjusting a loss function, ii) decoupling classifier learning, or iii) proposing a new multi-head architecture called experts. In this paper, we tackle the problem from a different perspective to augment a training dataset to enhance the sample diversity of minority classes. Specifically, our method, namely Confusion-Pairing Mixup (CP-Mix), estimates the confusion distribution of the model and handles the data deficiency problem by augmenting samples from confusion pairs in real-time. In this way, CP-Mix trains the model to mitigate its weakness and distinguish a pair of classes it frequently misclassifies. In addition, CP-Mix utilizes a novel mixup formulation to handle the bias in decision boundaries that originated from the imbalanced dataset. Extensive experiments demonstrate that CP-Mix outperforms existing methods for long-tailed image recognition and successfully relieves the confusion of the classifier.
\end{abstract}


\section{Introduction}
\label{section:introduction}

Large-scale datasets~\cite{imagenet, imagenet21k}, algorithmic innovations~\cite{adam, ioffe2015batch}, and powerful network architectures~\cite{resnet, dosovitskiy2020image} have driven remarkable achievements in computer vision over recent decades. However, these advancements primarily rely on clean, well-curated datasets like CIFAR-100 or ImageNet, which are balanced with a comparable amount of samples for each image class. In contrast, real-world data collection processes often result in biased and imbalanced datasets with a long-tail distribution. This distribution is characterized by a large number of minority classes containing very few samples, while the majority of samples belong to a small number of majority classes. Consequently, the standard training strategy based on empirical risk minimization struggles to generalize effectively to minority classes, leading to poor performance in these categories.

Given the prevalence of long-tailed data in practical scenarios, numerous research efforts have been directed toward addressing long-tailed problems. An intuitive approach to addressing this issue is data augmentation, as the key challenge is the lack of samples in minority classes. However, implementing data augmentation is not straightforward in such contexts due to the extreme scarcity of minority samples. Often, the majority classes benefit more from the augmentation method, further enlarging the discrepancy between majority and minority classes. To address these challenges, researchers have explored simper methods by synthesizing minority data points, using techniques such as SMOTE and its variations~\cite{smote, smote2, m2m}. One such approach is \emph{`Mixup'}, which linearly interpolates two existing data points to create a new data point~\cite{mixup}. Despite its conceptual simplicity, Mixup has proven effective in improving generalization~\cite{manifold, cutmix, Augmix, comix, puzzlemix, genlabel, saliencymix, gbmix} and long-tailed learning~\cite{remix, mixboost, mislas, cmo, csa, otmix}.

In this paper, we focus on effectively applying Mixup in scenarios with a large number of severely imbalanced classes. We first demonstrate that Mixup does not improve prediction accuracy for minority groups when extreme imbalances exist in the data distribution. Instead, it amplifies the performance gap between majority and minority groups because Mixup does not consider class imbalance or relationships among different classes. Vanilla Mixup randomly samples two data points from the training set, which becomes increasingly ineffective for imbalanced and complex datasets due to the difficulty of sampling minority classes and pairing them with confusing classes. To address this, we developed the confusion-pairing mixup (CP-Mix) method, which specifically samples from \emph{`confusion pairs'} that are prone to misclassification. With this modification, we significantly improve minority accuracy, even when the imbalance factor is as high as $996$, e.g., for Places-LT. In vanilla Mixup, samples from minority groups are rarely used and often paired with unrelated classes. Our method focuses on minority classes, pairing them with useful samples to enhance model generalization. Through an ablation study, we demonstrate that strategic sampling from confusion pairs is the primary contributor to our method's improved performance.

While many strategies address the long-tailed problem, we believe that our proposed CP-Mix is uniquely strong as a simple yet powerful tool. First, it provides better generalization than approaches based on re-weighting or re-sampling~\cite{reweighting, undersampling, reweight1, reweight2, oversampling, cbl} or modifying the objective function~\cite{focal, ldam, difficulty1, eql, lade, vstla}, which often lead to overfitting and memorization of minority samples. Secondly, CP-Mix requires significantly less computation than solutions based on training multiple-experts~\cite{bbn, lfme, ride, ace, ncl, sade}. Finally, it is easy to implement on top of any existing learning algorithm, being model-agnostic.

\begin{figure}[t!]
    \centering
    \begin{subfigure}[b]{0.55\textwidth}
        \centering
        \includegraphics[width=0.85\textwidth]{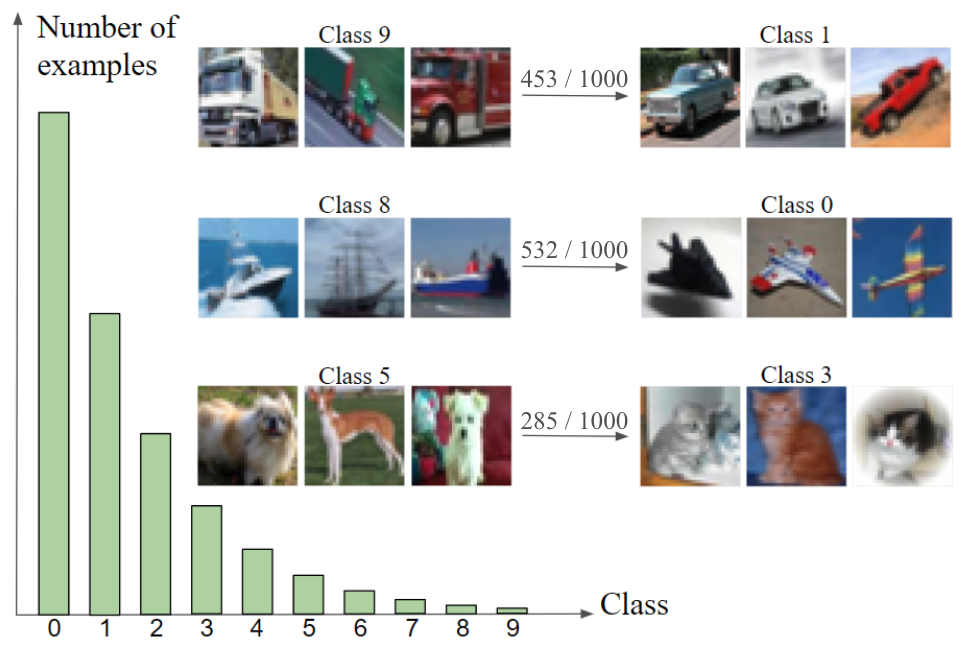}
        \subcaption{Class distribution of an imbalanced training set CIFAR10-LT-200 and the model's misclassification tendency.}
        \label{figure:confusion_tendency}
    \end{subfigure}
    \hfill
    \begin{subfigure}[b]{0.44\textwidth}
        \centering
        \includegraphics[width=0.85\textwidth]{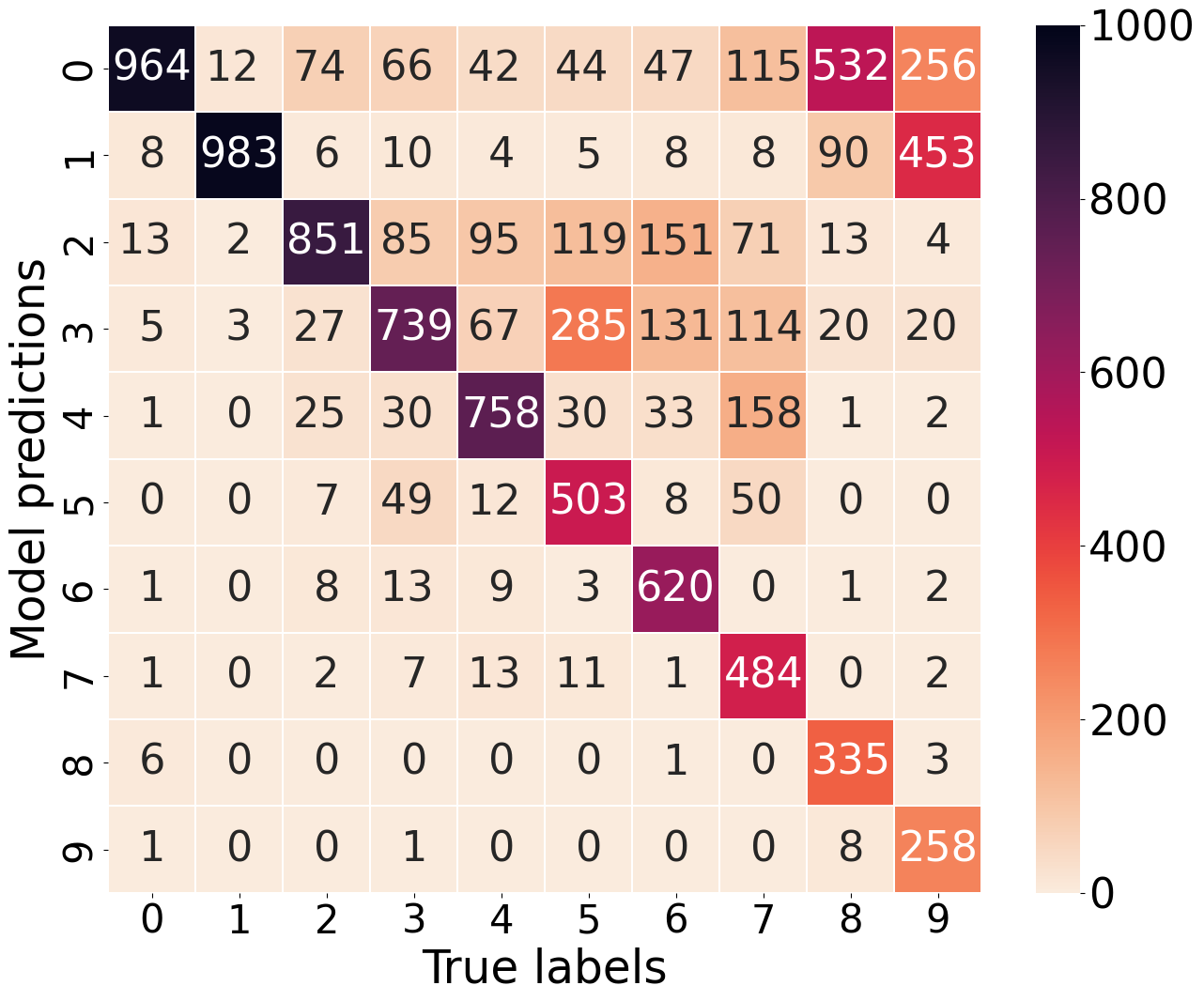}
        \caption{Confusion matrix of the classifier trained using CIFAR10-LT-200 dataset.}
        \label{figure:confusion_cifar10-lt-200_ce}
    \end{subfigure}
    \caption{
    (a) The model's tendency to misclassify the examples in the minorities (truck, ship, and dog) to their similar majorities (car, plane, and cat).
    \textbf{The model captures the similarities between classes and wrongly exploits its capacity.}
    (b) The confusion matrix of the model trained using the imbalanced dataset.
    $x$ and $y$ axis denote true labels and predictions, respectively, and each number in a cell denotes the number of predictions.
    \textbf{There are clear relationships between semantically similar classes.}
    }
    \label{figure:confusion_tendency_and_confusion}
\end{figure}

To summarize our contributions:
\begin{itemize}
    \item We identify a novel problem that vanilla mixup strategies exhibit a bias-amplifying phenomenon under very-long-tailed regimes where we have a large number of classes and the imbalance is significant between majority and minority classes.
    \item We propose CP-Mix algorithm that combines ideas of confusion pair sampling and class-imbalance-aware mixing to achieve more targeted intervention on the most vulnerable pairs while achieving generalization. 
    \item We provide through experimental evaluation that CP-Mix achieves improvements over baselines in various long-tailed datasets.
\end{itemize}

This paper is organized as follows. Section \ref{section:background} introduces the background of long-tailed recognition and mixup augmentation. Then, we discuss how imbalance in the training set affects the models and how mixup cannot improve them in Section \ref{section:confusion}. CP-Mix is described in Section \ref{section:cpmix}, and its superiority is demonstrated in Section \ref{section:experiments} through extensive experiments.

\section{Background and Related Works}
\label{section:background}

\subsection{Long-tailed image classification}
\label{subsection:background-longtailed}

Long-tailed recognition aims to train a model that performs well given a training dataset with a long-tailed distribution. We target an image classification task that the class distribution is severely imbalanced with $C$ classes. Let the training dataset ${\cal D}_s = \{(x_i, y_i)\}^{N-1}_{i=0}$ consists of $N$ training points where $x_i$ and $y_i$ denote an input and its corresponding class label, sampled from the input space $\cal X$ and the label set ${\cal C} = \{0, \cdots , C-1\}$, respectively. ${\cal D}_s$ is an union of $C$ disjoint subsets ${\cal D}_c$ for $c \in {\cal C}$ where ${\cal D}_c = \{ (x_i, y_i) | (x_i, y_i) \in {\cal D}_s, y_i = c \}$. We denote the cardinality of a subset ${\cal D}_c$ as $n_c$ where $\sum^{C-1}_{c=0} n_c = N$ and $n_0 > n_1 > \dots > n_{C-1}$ without loss of generality. We also denote an imbalance factor of ${\cal D}_s$ as $\rho = n_0 / n_{C-1} \gg 1$.

We term $f_\theta : {\cal X} \to {\mathbb R}^C$ as a model parameterized by learnable parameters $\theta$ that maps an input to a label and a prediction as $\hat{y}_\theta(x) = \arg \max f_\theta(x)$. With a bit of abuse of notations, we use $y$ to denote both the class label and its one-hot vector. The ERM objective is denoted by 
\begin{equation}\label{equation:erm-objective}
    {\mathcal L}_{ERM}(\theta, l) 
    = {\mathbb E}_{(x, y) \sim {\cal P}}[l(f_\theta (x), y)]
    \approx {1 \over N} \sum_{(x_i, y_i) \in {\cal D}_s} l(f_\theta (x_i), y_i),
\end{equation}
where $\cal P$ and $l(f_\theta (x), y)$ are the data distribution and sample-wise classification loss, respectively.

\subsection{Mixup}
\label{subsection:mixup}

Mixup \cite{mixup} is a simple and effective data augmentation strategy that linearly interpolates samples $(x_1, y_1)$ and $(x_2, y_2)$ to improve the generalization capacity of the model. Using a linear interpolation function $M(d_1, d_2, \lambda) = \lambda \times d_1 + (1 - \lambda) \times d_2$ for internal division where $d$ is input or label and $\lambda \in [0, 1]$, mixup formulation can be written as
\begin{equation}\label{equation:vanillamix_x}
    \tilde{x}_{mix} = \lambda \times x_1 + (1 - \lambda) \times x_2 = M(x_1, x_2, \lambda),
\end{equation}
\begin{equation}\label{equation:vanillamix_y}
    \tilde{y}_{mix} = \lambda \times y_1 + (1 - \lambda) \times y_2 = M(y_1, y_2, \lambda),
\end{equation}
where $y_1$ and $y_2$ are one-hot vectors of labels, and $\lambda$ is sampled from a random distribution. Based on theoretical analysis and symmetry of data distribution, the authors of \cite{mixup} proposed to use Beta distribution to sample $\lambda \sim Beta(\alpha, \alpha)$. To avoid confusion, we denote $\lambda$ and $\alpha$ as a mixing parameter and mixup hyperparameter, respectively.

Mixup uses $(\tilde{x}_{mix}, \tilde{y}_{mix})$ instead of real data points to train the model. The learning objective is
\begin{equation}\label{equation:mixup-objective}
    {\cal L}_{Mixup}(\theta, l) 
    = {\mathbb E}_{(\tilde{x}, \tilde{y}) \sim \tilde{{\cal P}}}[l(f_\theta (\tilde{x}), \tilde{y})] 
    \approx {1 \over N'} \sum_{(\tilde{x}_i, \tilde{y}_i) \in {\tilde{\cal D}_s}} l(f_\theta (\tilde{x}_i), \tilde{y}_i),
\end{equation}
where $\tilde{\cal P}$ and $\tilde{\cal D}_s$ are the distribution and dataset with cardinality $N'$ of mixed samples $(\tilde{x}_{mix}, \tilde{y}_{mix})$.

With negligible additional computation, mixup converts the empirical risk minimization (ERM) into the vicinal risk minimization (VRM) \cite{vrm}. Inspired by its simplicity, many extensions have been proposed to interpolate samples in manifold space \cite{manifold}, randomly crop a patch in an image and replace it with another patch from another image \cite{cutmix}, utilize self-augmentation to generate new samples \cite{Augmix} and use a saliency map to augment contextual information \cite{comix, puzzlemix, saliencymix}. Also, theoretical aspects of mixup are considered in many works \cite{mixup-datadependency, adamixup, unimixup, howdoesmixup}.

Recently, RegMixup \cite{regmixup} has been proposed to use a mixup objective as a regularization for VRM instead of a sole training objective.
Based on a thorough analysis, RegMixup argues that combining ERM and mixup objectives improves generalization and out-of-distribution robustness simultaneously.
In this paper, we combine the mixup objective revised for the long-tailed training set with the ERM objective.

\subsubsection{Mixup for long-tailed image recognition}
\label{subsubsection:background-mixupforlongtailed}

Several works have tried to handle long-tailed image classification using Mixup. Remix \cite{remix} considers the imbalance ratio to determine the mixed labels. The iterative algorithm is used in Mixboost \cite{mixboost} to synthesize additional samples in imbalanced datasets. The authors of \cite{bagoftricks} use mixup as a trick to boost long-tailed visual recognition. Also, authors of \cite{unimixup, mislas} solve the problem of mis-calibrated model using mixup in the long-tailed distribution. CMO \cite{cmo} uses the Cutmix \cite{cutmix} to synthesize new samples from the background of majorities and instances of minorities. CSA \cite{csa} effectively augments the training samples by removing irrelevant contexts. The most related work to ours is OTMix \cite{otmix} which learns the relationships between classes and utilizes CutMix for augmentation. Unlike previous works, we propose a novel approach to regularize the model by mixing samples from confusing classes estimated from the training set in real-time to compensate for the weak points.

\section{Confusion Tendency under Long-Tailed Distribution}
\label{section:confusion}

\begin{figure}[t!]
    \centering
    \begin{subfigure}[b]{0.17\textwidth}
        \centering
        \includegraphics[width=1.1\textwidth]{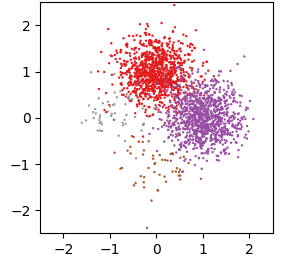}
        \subcaption{Train data.}
        \label{figure:toys/train}
    \end{subfigure}
    \hfill
    \begin{subfigure}[b]{0.17\textwidth}
        \centering
        \includegraphics[width=1.1\textwidth]{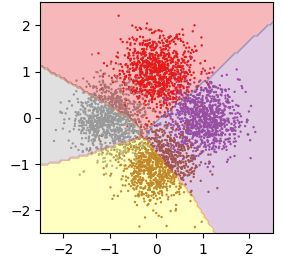}
        \subcaption{ERM.}
        \label{figure:toys/erm}
    \end{subfigure}
    \hfill
    \begin{subfigure}[b]{0.17\textwidth}
        \centering
        \includegraphics[width=1.1\textwidth]{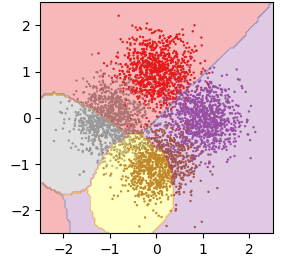}
        \subcaption{Mixup.}
        \label{figure:toys/mixup}
    \end{subfigure}
    \hfill
    \begin{subfigure}[b]{0.17\textwidth}
        \centering
        \includegraphics[width=1.1\textwidth]{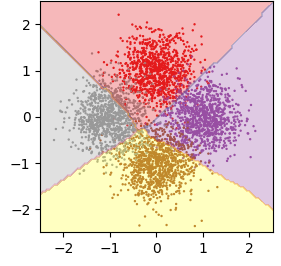}
        \subcaption{ERM with reg.}
        \label{figure:toys/pairreg}
    \end{subfigure}
    \hfill
    \begin{subfigure}[b]{0.24\textwidth}
        \centering
        \includegraphics[width=1.0\textwidth]{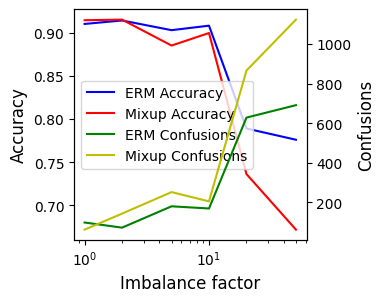}
        \subcaption{Acc and confusions.}
        \label{figure:toys/plot}
    \end{subfigure}
    \caption{
    (a) 2D 4-way toy example with 1000 and 50 samples in majority and minority classes.
    The majority classes red and purple are similar to minorities grey and yellow, respectively.
    (b-d) Decision boundaries of the classifiers on a balanced test dataset.
    (b, c) \textbf{The ERM classifier has a biased boundary toward minorities, and the Mixup classifier has a more restricted region for minorities since Mixup mainly occurs between and within red and purple points.}
    (d) The ERM classifier regularized by a mixup objective where similar majority and minority classes are mixed.
    \textbf{Mixup occurs between red and grey points or purple and yellow points for the regularization.
    It successfully improves the decision boundary while maintaining the structure of the boundary.}
    (e) Accuracy and confusion of ERM classifier as imbalance factors vary.
    Confusion denotes the sum of confusion values between two pairs of adjacent majority and minority, grey points in red region and yellow points in purple region.
    \textbf{Although Mixup improves generalization in small imbalances, it fails as the imbalance factor increases.}
    }
    \label{figure:toys}
\end{figure}

When a classifier is trained on an imbalanced training set, it struggles to distinguish similar classes with different numbers of samples in the training set. It tends to classify minority samples as the majority counterpart, resulting in inferior performance. It is an important and serious problem as it commonly occurs in real-world datasets but is not observable in balanced training.

We define confusion of the model $f_\theta$ from class $c_i$ to $c_j$ on dataset $\cal D$ as
\begin{equation}\label{equation:confusion}
    {\it{C}_{i, j}}(f_\theta, {\cal D}) = |\{(x, c_i, c_j) | f_\theta(x) = c_j, (x, c_i) \in {\cal D}\}|, 
\end{equation}
to evaluate how much the model $f_\theta$ confuses class $c_i$ as $c_j$. When the distribution of $C_{i, j}$ differs from the uniform as shown in Figure \ref{figure:confusion_tendency_and_confusion}\ref{figure:confusion_cifar10-lt-200_ce}, we can say the model has high confusion. Although Mixup can augment minority samples with negligible additional complexity, it has limitations in reducing high confusion $C_{i, j}$ in long-tailed distribution. In fact, Mixup focuses on the generalization of majorities for imbalanced training datasets, aggravating minority generalization. In this section, we investigate how mixup fails to resolve confusion and improve classification. Also, we conduct further analysis on confusion matrices to achieve a deeper understanding of long-tailed distribution\footnote{Confusions $C_{i, j}$ are non-diagonal values of confusion matrix.}.

\subsection{Toy Example}
\label{subsection:toy-example}

This subsection describes how the decision boundary of the classifier trained on the imbalanced set is biased using a 4-way classification dataset on a 2D plane, as shown in Figure \ref{figure:toys}\ref{figure:toys/train}. The classifier trained using the ERM objective has a biased decision boundary, as shown in Figure \ref{figure:toys}\ref{figure:toys/erm}, failing to achieve generalization in minorities. When we train a Mixup classifier on the same dataset, Mixup frequently occurs between and within red and purple majorities, providing negligible changes to the decision boundary between them. Instead, rare mixups containing minority samples make it harder for the classifier to learn the boundary surrounding minorities due to the randomness. Figure \ref{figure:toys}\ref{figure:toys/mixup} demonstrates how mixup destroys the structure of the decision boundary related to minorities when data imbalance is not considered.

In Figure \ref{figure:toys}\ref{figure:toys/plot}, we can observe the relationship between accuracy of ERM classifier and confusion $C_{i, j}$ as the imbalance factor $\rho$ differs in the toy example. Confusions on the right axis denote the sum of misclassifications from minority classes to their adjacent majorities, grey points in red region and yellow points in purple region. The increase of $\rho$ results in higher confusion between adjacent majority and minority and decreased accuracy. Moreover, the negative effect of mixup grows as there is a severe imbalance between majorities and minorities.

We regularized the ERM classifier using a mixup objective, where mixup occurs between similar majority and minority, to reduce the confusions. When the augmented samples between red and grey points, and purple and yellow points are used for regularization, the decision boundary is less biased, as shown in Figure \ref{figure:toys}\ref{figure:toys/pairreg}. It demonstrates that mixup considering imbalance and similarity of classes are critical for improving the classifiers.

\begin{figure}[t!]
    \centering
    \begin{subfigure}[b]{0.34\textwidth}
        \centering
        \includegraphics[width=0.95\textwidth]{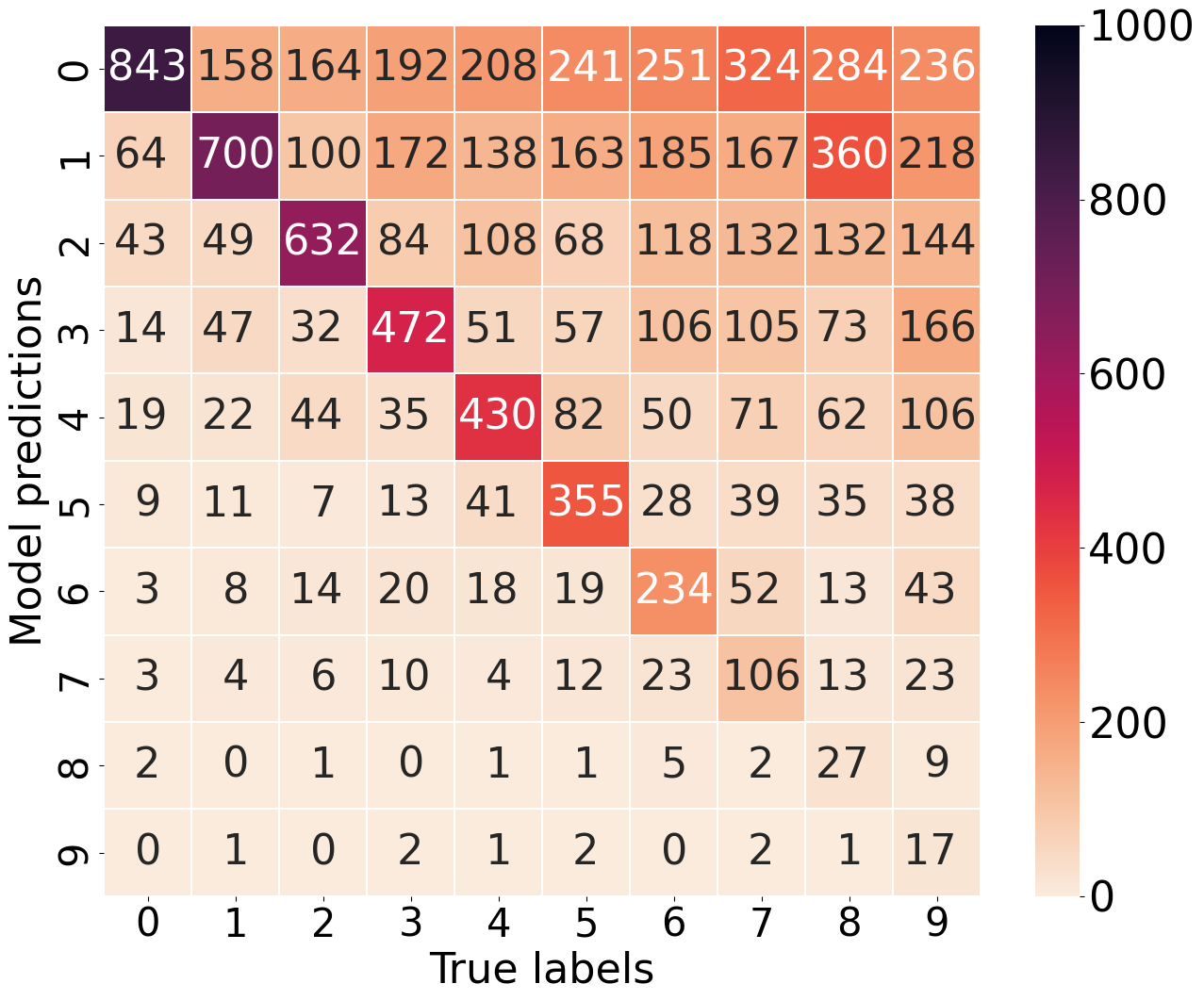}
        \subcaption{Confusion matrix of ERM classifier.}
        \label{figure:cifars/confusion-ce}
    \end{subfigure}
    \hfill
    \begin{subfigure}[b]{0.34\textwidth}
        \centering
        \includegraphics[width=0.95\textwidth]{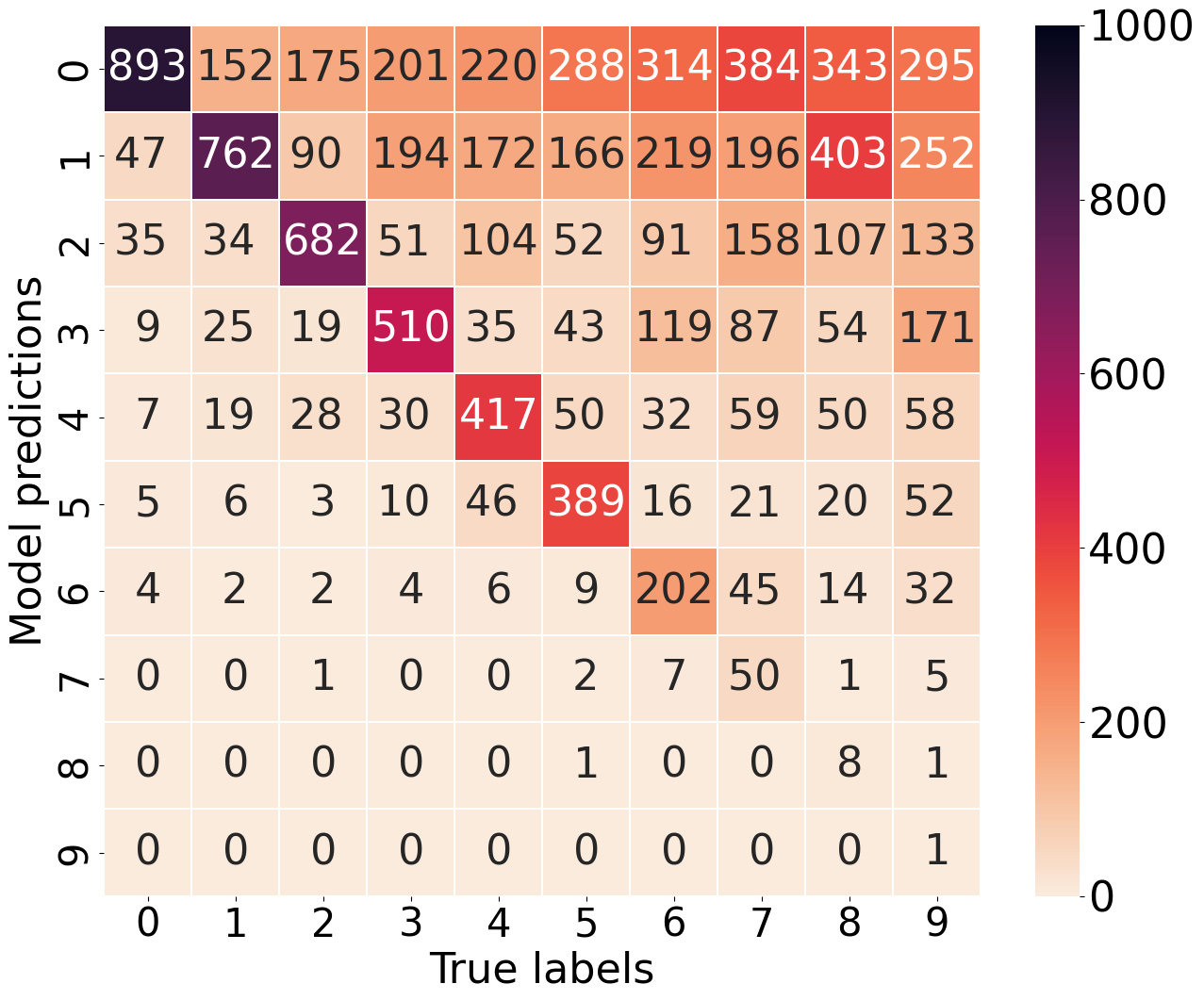}
        \subcaption{Confusion matrix of Mixup classifier.}
        \label{figure:cifars/confusion-mixup}
    \end{subfigure}
    \hfill
    \begin{subfigure}[b]{0.29\textwidth}
        \centering
        \includegraphics[width=0.94\textwidth]{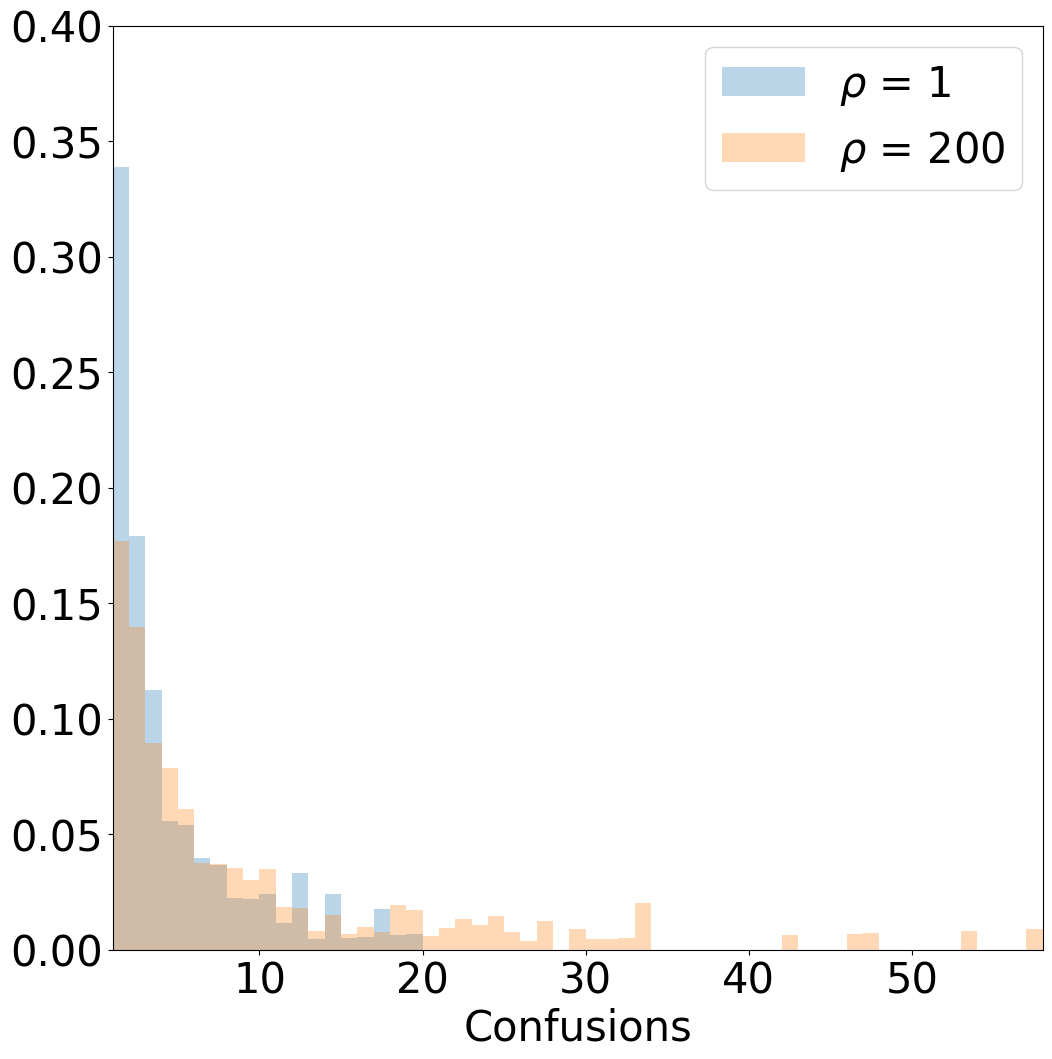}
        \subcaption{Histogram of confusions $C_{i. j}$.}
        \label{figure:cifars/histogram}
    \end{subfigure}
    \caption{
    (a, b) Confusion matrices of ERM and Mixup classifiers trained on CIFAR100-LT dataset, respectively. The imbalance factor is 200, and classes are grouped into 10 subgroups for better visualization. \textbf{Mixup exacerbates the confusion of model, and only improves the generalization of majorities.}
    (c) Histogram of confusion values between pairs of classes for balanced CIFAR100 and CIFAR100-LT-200 datasets. Among 100 samples in each class, the maximum confusion value between two classes increases from 20 to more than 50 as $\rho$ increases to $200$. 
    }
    \label{figure:cifars/confusion}
\end{figure}

\subsection{CIFAR-LT}
\label{subsection:cifar-example}

Similar problems are observed in classifiers trained in complex imbalanced datasets like CIFAR100-LT, as shown in Figure \ref{figure:cifars/confusion}. We consolidate the 100 classes into 10 new groups, combining 10 classes into each group to simplify visualization. ERM classifier suffers from severe confusion, as demonstrated in the upper right region of confusion matrix in Figure \ref{figure:cifars/confusion}\ref{figure:cifars/confusion-ce}. Most samples from minority classes are misclassified as majorities, while less than a hundred of ten thousand samples are predicted as the bottom 20 classes. Figure \ref{figure:cifars/confusion}\ref{figure:cifars/confusion-mixup} demonstrates that sparsity of the bottom rows becomes more severe with Mixup, where less than a hundred samples are classified as bottom 30 classes. It shows that Mixup improves the generalization of majority groups rather than minorities, aggravating the bias of the classifiers.

We also observe a similar tendency of uneven confusion values $C_{i, j}$ for the ERM classifiers trained on CIFAR100-LT datasets. Figure \ref{figure:cifars/confusion}\ref{figure:cifars/histogram} demonstrates the difference in confusion values between balanced and imbalanced datasets. While the maximum confusion value is less than 20 for the balanced CIFAR100 dataset, it increases to more than 50 out of 100 samples in a class as the imbalance factor becomes $200$. Since the distribution of confusion values is long-tailed, more focus on the tail of confusion distribution is required for better performance. As we consider larger datasets with vast numbers of classes, it becomes more critical to concentrate specific pairs of classes in the tail distribution for Mixup since each pair is less likely to be sampled pairwise in random sampling.

\section{Confusion-Pairing Mixup}
\label{section:cpmix}

We propose Confusion-Pairing Mixup (CP-Mix), which guides the model to classify confusing classes to improve minority generalization using mixup. Minority samples tend to be misclassified into some specific majority classes depending on the context and relation of classes when the train distribution is imbalanced, as shown in Figure \ref{figure:confusion_tendency_and_confusion}. To handle this tendency, we propose estimating the model's confusion distribution using the predictions from the imbalanced training set in real-time. Based on this estimation, CP-Mix samples and mixes data points, considering imbalances and confusions between classes. In the following subsections, the components of CP-Mix are described in detail.

\subsection{Confusion-Pairs}
\label{subsection:cpmixup-confusionpairs}

CP-Mix estimates confusion distribution by collecting pairs of classes $(c_t, c_m)$ that the model $f_\theta$ misclassifies an example $x$ with a true label $c_t$ as $c_m$. We use the imbalanced training set as a source to collect confusion pairs, so they are collected during the training phase. We denote a multiset (or bag) of these pairs as confusion-pairs ${\cal CP}_e$ obtained from the current model at epoch $e$ as: 
\begin{equation}\label{equation:confusion-pairs}
    {{\cal CP}_e}({\cal D}_s) = [(c_t, c_m) | (x, c_t) \in {\cal D}_s, c_m = {\hat y}_\theta(x), c_t \not = c_m]
\end{equation}
where the size of ${\cal CP}_e$ is the number of misclassification at epoch $e$. We use square brackets to demonstrate a multiset to avoid confusion with a set. Multiset ${\cal CP}_e$ allows the duplication of pairs so that it can describe the frequency of pairs\footnote{It can be replaced by counting actual confusion values $C_{i, j}$, but we use the concept of multiset for easier understanding.}. These confusion-pairs are accumulated to ${\cal CP}$ so that the model can consider historical confusion tendencies. CP-Mix samples pairs from $\cal CP$ and then data points given these pairs, while other mixup methods randomly match points in a mini-batch to synthesize new points. With CP-Mix, the model can concentrate more on class pairs of misclassifications and be trained to correct the confusion.

\subsection{Mixing functions}
\label{subsection:cpmixup-mixingfunctions}

Using different mixing parameters $\lambda$ for input and label is a natural extension of vanilla mixup as below:
\begin{equation}\label{equation:decouplemix_x}
    \tilde{x}_{mix} = \lambda_x \times x_1 + (1 - \lambda_x) \times x_2 = M(x_1, x_2, \lambda_x),
\end{equation}
\begin{equation}\label{equation:decouplemix_y}
    \tilde{y}_{mix} = \lambda_y \times y_1 + (1 - \lambda_y) \times y_2 = M(y_1, y_2, \lambda_y).
\end{equation}
When the training dataset is severely imbalanced, the Mixup classifier learns a biased decision boundary, as shown in Figure \ref{figure:toys}\ref{figure:toys/mixup} since it has been exposed to more samples mixed from samples in majority classes. To compensate for this behavior, mixed labels more favorable to minorities are required for a better decision boundary.

We propose to consider the imbalance of the training set for the label mixing function in Equation~\eqref{equation:decouplemix_y}. Given a pair of examples $(x_1, y_1)$ and $(x_2,  y_2)$, our label mixing function considers the imbalance between two classes $y_1$ and $y_2$ as
\begin{equation}\label{equation:lambday}
    \lambda_y(\lambda) = t \times \lambda + (1 - t) \times {n_{y_2} \over {n_{y_1} + n_{y_2}}}
\end{equation}
where $t$ is a hyperparameter for label mixing and $\lambda$ is sampled from beta distribution. With this formulation, the output of label mixing can be written as:
\begin{equation}\label{equation:y_cp}
    \tilde{y}_{CP} 
    = M(y_1, y_2, \lambda_y (\lambda))
    = M(M(y_1, y_2, \lambda), M(y_1, y_2, {n_{y_2} \over {n_{y_1} + n_{y_2}}}), t).
\end{equation}
This is equivalent to interpolating two terms: the first term can be thought of as a randomly mixed label, and the second term is mixed with a label-imbalance-aware parameter. While the first term provides generalization, as in vanilla Mixup, the second term tackles the underlying label imbalance in the training set. Our unique approach departs from the method employed in Remix~\cite{remix}, which uses an imbalance ratio for a threshold to decide the mixing parameter, and aims to achieve generalization alongside label imbalance mitigation. We use a randomly sampled $\lambda$ for input mixing parameter, as in the vanilla Mixup.

\subsection{CP-Mix}
\label{subsection:cpmixup-cpmix}

If ${\cal CP}$ is used from the beginning of training, there exist two problems. First, minorities would be repeatedly used in the training, accelerating the overfitting on minorities. As a result, pairs including minorities would not be collected after a few epochs. Second, ${\cal CP}$ is not populated by pairs that demonstrate the similarities or relations of classes. If the classifier is trained to remedy its weakness in real time from the beginning, estimated confusion distribution will not reflect the underlying relationships of classes. To remedy these problems, we design CP-Mix training to follow ERM training, where ${\cal CP}$ is collected during the whole training time, but mixup regularization with confusion pairs works at the second stage. By doing so, we not only populate ${\cal CP}$ with pairs that the ERM classifier confuses but also reduce the computational overhead during training.

We construct a mini-batch for CP-Mix in the following steps given the multiset $\cal CP$, as shown in Algorithm \ref{algo:cpmix}. First, we sample a set of $B'$ classes $\{(c_{b, t})\}_{b=0}^{B'-1}$ uniformly from the label set ${\cal C} = \{0, \cdots , C-1\}$ where $B'$ is a batch size for mixup. This sampling does not consider class distribution in confusion-pairs $\cal CP$ since it is imbalanced due to the source training set's imbalance. Then we sample a pair of classes $(c_{b, t}, c_{b, m})$ for each $c_{b, t}$ where $c_{b, m}$ is sampled from a subset $\{(c_{t}, c_{m}) | (c_t, c_m) \in {\cal CP}, c_t = c_{b, t}\}$. After we obtain a set of $B'$ confusion-pairs $\{(c_{b, t}, c_{b, m})\}_{b=0}^{B'-1}$, we sample $x_{b,t}$ and $x_{b,m}$ from the subsets ${\cal D}_{c_{b, t}}$ and ${\cal D}_{c_{b, m}}$ for each confusion pair $(c_{b, t}, c_{b, m})$. Finally, we perform pairwise mixing of $x_{b,t}$ and $x_{b,m}$ using the mixing functions proposed in Section \ref{subsection:cpmixup-mixingfunctions}.

\begin{algorithm}[t]
\caption{CP-Mix}
\label{algo:cpmix}
\KwInput{$f_\theta$, ${\cal D}_s = \{(x_i, y_i)\}^{N-1}_{i=0}$}
\textbf{Hyperparameter:} $E, E_{cp}, B, B', \gamma$\; 
Initialize $f_\theta$, Confusion-pairs ${\cal CP} \gets []$\; 

\For{$e = 1$ \KwTo $E$}{
    \ForEach{batch ${\cal B} = \{(x_i, y_i)\}^{B-1}_{i=0} \in {\cal D}_s$}{
        ${\cal L}(\theta) \gets {\cal L}_{ERM}(\theta, l_{ERM})$ using ${\cal B}$\;
        Get predictions $P = \{(x_i, y_i, y_{i, p})\}^{B-1}_{i=0}$\;
        Append misclassifications $[(y_i, y_{i, p}) \mid (x_i, y_i, y_{i, p}) \in P, y_i \neq y_{i, p}]$ to ${\cal CP}$\;
        
        \If{$e > E_{cp}$}{
            Sample classes $\{(c_{b, t})\}_{b=0}^{B'-1}$ uniformly from ${\cal C}$\;
            Sample confusion-pairs $\{(c_{b, t}, c_{b, m})\}_{b=0}^{B'-1}$ from ${\cal CP}$ given $\{(c_{b, t})\}_{b=0}^{B'-1}$\;
            Sample batch ${\cal B}_{CP} = \{((x_{b, t}, c_{b, t}), (x_{b, m}, c_{b, m}))\}_{b=0}^{B'-1}$ from ${\cal D}_s$\;
            Mix samples in ${\cal B}_{CP}$ and compute ${\cal L}_{CP}(\theta, l_{CP})$\;
            ${\cal L}(\theta) \gets {\cal L}(\theta) + \gamma \times {\cal L}_{CP}(\theta, l_{CP})$\;
        }
        
        Update $\theta$ using ${\cal L}(\theta)$\;
    }
}
\KwOutput{$f_\theta$}
\end{algorithm}

Combining the above proposals, CP-Mix objective is
\begin{equation}\label{equation:cp-objective}
    {\cal L}_{CP}(\theta, l) 
    = {\mathbb E}_{(\tilde{x}, \tilde{y}) \sim \tilde{\cal P}_{CP}}[l(f_\theta (\tilde{x}), \tilde{y})]
    \approx {1 \over N''} \sum_{(\tilde{x}_i, \tilde{y}_i) \in {\tilde{\cal D}_{s, CP}}} l(f_\theta (\tilde{x}_i), \tilde{y}_i),
\end{equation}
where $\tilde{\cal P}_{CP}$ is a CP-mixed data distribution and $\tilde{\cal D}_{s, CP}$ is a set of CP-mixed examples with $N''$ cardinality. As discussed, a mini-batch ${\cal B}_{CP}$ for mixup is sampled based on confusion-pairs.

We obtain the overall objective as
\begin{equation}\label{equation:overall-objective}
    {\cal L}(\theta) = {\cal L}_{ERM}(\theta, l_{ERM}) + \gamma {\cal L}_{CP} (\theta, l_{CP}),
\end{equation}
where $\gamma$ is a hyper-parameter. We use balanced softmax \cite{bs} for $l_{ERM}$ and cross-entropy loss for $l_{CP}$, considering underlying data distribution for these objectives.\footnote{The data distribution for $l_{CP}$ already considers the imbalance and model's bias, while the data distribution for $l_{ERM}$ is imbalanced.} In practice, we apply additional mixup regularization with cross-entropy loss using the imbalanced training set to stabilize the training. More details can be found in Appendix~\ref{appendix:objective}.

\section{Experiments}
\label{section:experiments}

In this section, we first provide the experimental settings to evaluate how CP-Mix works when the training set is severely imbalanced. Then, we demonstrate that CP-Mix outperforms existing state-of-the-art methods across multiple long-tailed datasets. This is followed by a detailed ablation study that dissects the various components contributing to the performance improvement of CP-Mix.

\subsection{Experimental settings}
\label{subsection:experiments-settings}

\paragraph{Datasets and model architecture} 
We use five datasets commonly benchmarked for long-tailed image recognition ({\it{i.e.}}, two CIFAR-LT \cite{cbl}, ImageNet-LT \cite{oltr}, Places-LT \cite{oltr}, and iNaturalist 2018 \cite{inaturalist}). CIFAR-LT are artificial versions of CIFAR10 and CIFAR100 datasets \cite{cifar} with exponential class imbalance factors varying from 10 to 200. ImageNet-LT and Places-LT are long-tailed versions of real-world datasets ImageNet \cite{imagenet} and Places \cite{place}, and we follow the standard procedure to build these datasets. iNaturalist 2018 is a large-scale real-world dataset with an imbalanced distribution. For a fair comparison, we follow previous works about network architecture. More information about the datasets and model architectures is summarized in Appendix~\ref{appendix:implementation:dataset}.

\paragraph{Evaluation protocols}
We use the uniform test set to evaluate methods and report the top-1 accuracy. Following \cite{oltr}, we also report the accuracy evaluated on three disjoint class subgroups divided according to the number of training examples in each class ({\it{i.e.}} many (more than 100), medium (from 20 to 100), and few (less than 20) shots) for large-scale datasets in Appendix~\ref{appendix:experiment}.

\paragraph{Baselines}
We compare CP-Mix with various representative approaches for long-tailed image classification. They are categorized into \textbf{a) Loss modification} (Balanced Softmax \cite{bs} and LADE \cite{lade}, including Cross Entropy as baseline), \textbf{b) Two-stage methods} (LDAM-DRW \cite{ldam} and MiSLAS \cite{mislas}), and \textbf{c) Mixup and generation methods} (Remix~\cite{remix}, CMO~\cite{cmo}, UniMix~\cite{unimixup} and OTmix~\cite{otmix}). We also include \textbf{d) Ensemble-based methods} (RIDE~\cite{ride}, ACE~\cite{ace}, and SADE~\cite{sade}) to evaluate the performance of CP-Mix combined with RIDE.

\paragraph{Implementation details}
We follow the common practices for implementation details from the literature. We trained the models using SGD optimizer with momentum 0.9. The learning rate for the training is 0.1, except for the Places-LT, which uses 0.01. We finetune the models using the class-balanced training dataset following the method proposed in \cite{lpft}. More details about training and hyper-parameters for CP-Mix can be found in Appendix~\ref{appendix:implementation:training}.

\begin{table}
    \centering
    \scalebox{1.0}{
        \begin{tabular}{c|cccc|cccc}
            \toprule
            \multicolumn{1}{c|}{Dataset} & \multicolumn{4}{c|}{CIFAR100-LT} & \multicolumn{4}{c}{CIFAR10-LT} \\
            \midrule
            Imbalance factor $\rho$ & $200$ & $100$ & $50$ & $10$ & $200$ & $100$ & $50$ & $10$ \\
            \midrule
            Cross Entropy$^\dagger$           & $34.84$ & $38.32$ & $43.26$ & $57.00$ & $65.68$ & $70.36$ & $77.10$ & $86.89$ \\
            LDAM-DRW$^\dagger$                & $39.03$ & $43.64$ & $47.31$ & $56.90$ & $76.14$ & $79.58$ & $82.96$ & $87.48$ \\
            Balanced Softmax$^\dagger$        & $38.01$ & $42.71$ & $46.68$ & $58.34$ & $74.48$ & $77.20$ & $81.26$ & $87.46$ \\
            MiSLAS                            & $  -  $ & $47.0$  & $52.3$  & $63.2$  & $  -  $ & $82.1$  & $85.7$  & $90.0$  \\
            Remix                             & $36.99$ & $41.94$ & $46.21$ & $59.36$ & $  -  $ & $79.76$ & $  -  $ & $89.02$ \\
            CMO                               & $  -  $ & $43.9$  & $48.3$  & $59.5$  & $  -  $ & $  -  $ & $  -  $ & $  -  $ \\
            UniMix                            & $42.07$ & $45.45$ & $51.11$ & $61.25$ & $78.48$ & $82.75$ & $84.32$ & $89.66$ \\
            OTmix (BS)                        & $  -  $ & $46.8$  & $52.3$  & $62.3$  & $  -  $ & $84.0$  & $86.5$  & $90.2$  \\
            \midrule
            ACE                               & $  -  $ & $49.6$  & $51.9$  & $  -  $ & $  -  $ & $81.4$  & $84.9$  & $  -  $ \\
            RIDE$^\dagger$                    & $46.34$ & $49.83$ & $52.73$ & $59.30$ & $79.16$ & $82.31$ & $84.61$ & $87.19$ \\
            CMO+RIDE                          & $  -  $ & $50.0$  & $53.0$  & $60.2$  & $  -  $ & $  -  $ & $  -  $ & $  -  $ \\
            SADE                              & $  -  $ & $49.8$  & $53.9$  & $63.6$  & $  -  $ & $  -  $ & $  -  $ & $  -  $ \\
            OTMix+RIDE                        & $  -  $ & $50.7$  & $53.8$  & $60.8$  & $  -  $ & $82.7$  & $85.2$  & $88.7$  \\
            \midrule
            CP-Mix$^\dagger$                  & $43.56$ & $48.20$ & $52.12$ & $61.91$ & $78.34$ & $82.44$ & $85.08$ & $89.87$ \\
            CP-Mix + RIDE$^\dagger$           & $47.37$ & $51.40$ & $55.77$ & $62.17$ & $79.67$ & $81.56$ & $84.23$ & $86.70$ \\
            \bottomrule
        \end{tabular}
    }
    \caption{
    Experimental results for CIFAR-LT datasets. $\dagger$ denotes results reproduced by us, and others are from the original papers. \textbf{The advantages of the proposed CP-Mix are highlighted as the task becomes challenging, especially for CIFAR100-LT with large imbalance factors.} CP-Mix is compatible with ensemble-based methods, outperforming other ensemble-based methods for challenging tasks.
    }
    \label{table:cifar}
\end{table}

\begin{table}
    \centering
    \scalebox{1.0}{
        \begin{tabular}{c|c|c|c}
            \toprule
            Dataset & ImageNet-LT & Places-LT & iNaturalist 2018 \\
            \midrule
            Cross Entropy$^\dagger$       & $45.5$ & $32.3$ & $64.5$ \\
            Balanced Softmax$^\dagger$    & $50.6$ & $39.3$ & $69.9$ \\
            LADE                          & $ - $  & $39.2$ & $69.3$ \\
            LDAM-DRW$^\dagger$            & $50.3$ & $ - $  & $66.9$ \\
            MiSLAS                        & $52.7$ & $40.4$ & $71.6$ \\
            Remix                         & $48.6$ & $ - $  & $70.5$ \\
            UniMix                        & $48.6$ & $ - $  & $70.5$ \\
            OTmix (BS)                    & $55.6$ & $ - $  & $71.5$ \\
            \midrule
            ACE                           & $54.7$ & $ - $  & $72.9$ \\
            RIDE$^\dagger$                & $55.4$ & $40.3$ & $72.6$ \\
            CMO+RIDE                      & $56.2$ & $ - $  & $72.8$ \\
            SADE                          & $ - $  & $40.9$ & $72.9$ \\
            OTmix+RIDE                    & $57.3$ & $ - $  & $73.0$ \\
            \midrule
            CP-Mix                        & $53.9$ & $42.3$ & $72.3$ \\
            CP-Mix+RIDE                   & $56.6$ & $ - $  & $73.7$ \\
            \bottomrule
        \end{tabular}
    }
    \caption{
    Results for large-scale datasets. $\dagger$ denotes results reproduced by us, and others are from the original papers. CP-Mix performs better when the number of categories and imbalance factor increase (iNaturalist 2018), demonstrating its superiority in tackling challenging problems.
    }
    \label{table:largescale}
\end{table}

\subsection{Long-tailed image classification}
\label{subsection:experiments-longtailed}

We evaluate CP-Mix on famous long-tailed classification datasets (CIFAR-LT with varying imbalance factors from $200$ to $10$, and three large-scale datasets). Experimental results for CIFAR-LT and large-scale datasets are reported in Table~\ref{table:cifar} and \ref{table:largescale}, respectively. The superiority of CP-Mix is highlighted for the complex tasks like CIFAR100-LT with imbalance factors of $200$ and $100$ or large-scale datasets while achieving comparable performance on smaller or less-imbalanced datasets. We also augment CP-Mix with RIDE to demonstrate its compatibility with ensemble-based method, and it outperforms other ensemble-based methods (ACE, RIDE and SADE). Notably, ensemble-based methods cannot outperform simple methods when imbalance ratio is not significant or the number of categories is not large. Additional results on three disjoint subgroups of categories are reported in Appendix~\ref{appendix:experiment:results}. CP-Mix is better at achieving trade-offs between sub-group performances. This implies that mixup is ineffective for large-scale images if we do not consider the imbalance and confusion.

OTmix~\cite{otmix} is the most relevant work that combines mixup with a confusion matrix. While it utilizes mixup to augment its proposed method based on optimal transport, CP-Mix has an important difference from OTmix. CP-Mix is carefully designed to reveal the model's confusion in real-time without having a balanced validation set, which is more practical in a real-world scenario. However, OTmix uses the balanced validation set to obtain the confusions of the model. This implies that our CP-Mix is more practical in tackling long-tailed problems.

\subsection{Analysis of CP-Mix}
\label{subsection:experiments-analysis}

We conduct further analysis to understand the advantages of CP-Mix. Our experiments aim to answer the following questions:
1) Can CP-Mix alleviate confusion-pairs? 
2) How does it improve class-wise performance? and
3) What compositions of CP-Mix boost long-tailed recognition?
All study in this subsection is conducted on the classifiers trained on CIFAR100-LT dataset with the imbalance factor $\rho = 200$ if not specified.

\subsubsection{Confusion matrix}
\label{subsubsection:experiments-analysis-confusion_matrix}

\begin{figure}[t]
    \centering
    \begin{subfigure}[b]{0.34\textwidth}
        \centering
        \includegraphics[width=1.0\textwidth]{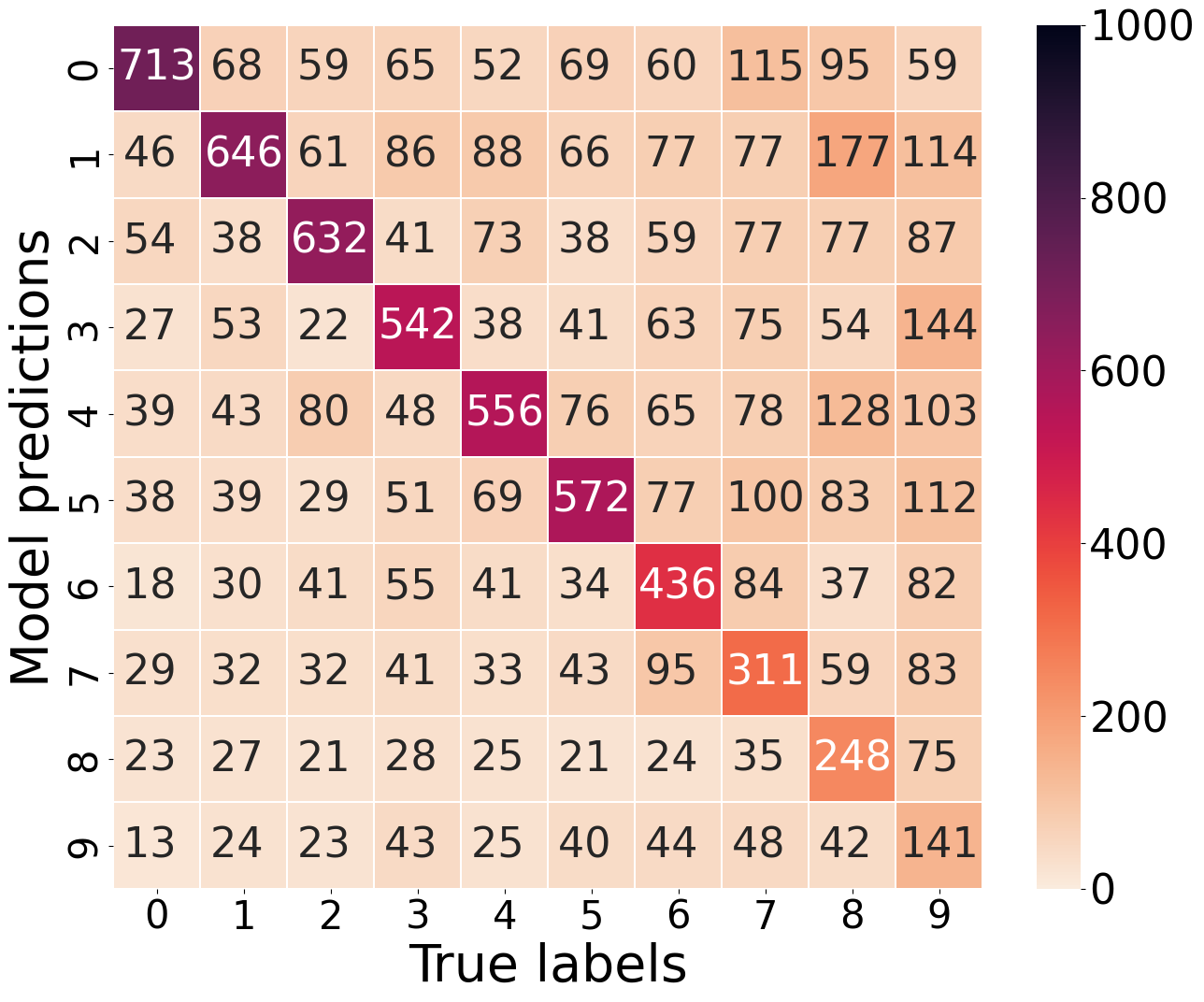}
        \subcaption{
        Confusion matrix of CP-Mix classifier on CIFAR100-LT-200 dataset.
        }
        \label{figure:analysis/confusion-matrix-cpmix}
    \end{subfigure}
    \hfill
    \begin{subfigure}[b]{0.31\textwidth}
        \centering
        \includegraphics[width=1.0\textwidth]{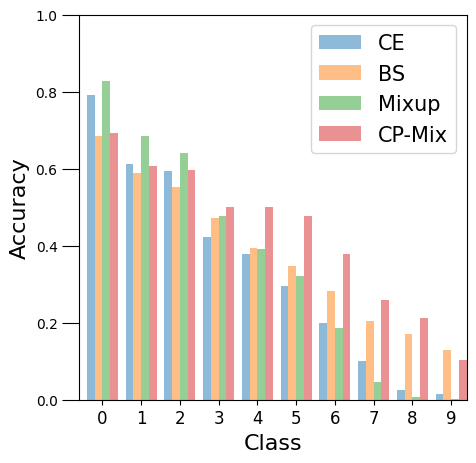}
        \subcaption{
        Class-wise accuracy on CIFAR100-LT-200 dataset.
        }
        \label{figure:analysis/comparison-acc-cifar100}
    \end{subfigure}
    \hfill
    \begin{subfigure}[b]{0.31\textwidth}
        \centering
        \includegraphics[width=1.0\textwidth]{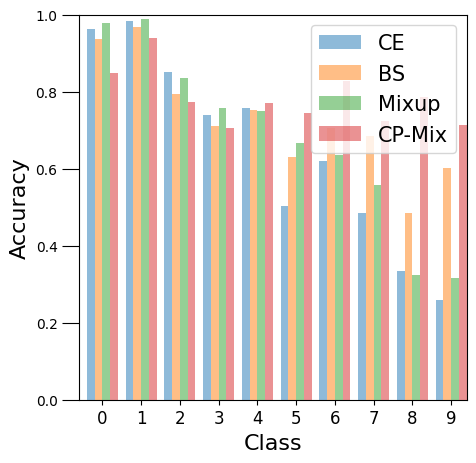}
        \subcaption{
        Class-wise accuracy on CIFAR10-LT200 dataset.
        }
        \label{figure:analysis/comparison-acc-cifar10}
    \end{subfigure}
    \caption{
    (a) Confusion matrix of the CP-Mix classifier trained on CIFAR100-LT-200 dataset. \textbf{It successfully reduces high confusion in the upper right region by sacrificing low confusion in the lower left region, which indicates the number of majorities misclassified as minorities.} This results in more balanced accuracies among categories.
    (b, c) Class-wise accuracy of the classifiers trained on CIFAR100-LT-200 and CIFAR10-LT-200 datasets. \textbf{CP-Mix significantly improves the accuracies on minority classes, resulting in more balanced sub-group accuracies.}
    }
    \label{figure:analysis}
\end{figure}

CP-Mix utilizes the confusion-pairs as a clue of the model's weakness and augments samples between these pairs to relieve confusions. We compare the confusion matrix of CP-Mix to that of ERM and Mixup classifiers to demonstrate that CP-Mix successfully relieves confusions of the model while Mixup cannot. Figure~\ref{figure:analysis}\ref{figure:analysis/confusion-matrix-cpmix} illustrates the model's weakness in misclassifying minorities to similar counterparts, showing its enhanced generalization to minority classes. On the other hand, confusion matrices in Figure~\ref{figure:cifars/confusion} demonstrate the clear tendency between the true classes and mispredictions. This implies that CP-Mix can capture the confusion distribution of the classifier, and sampling from the confusion-pairs is beneficial in relieving the confusion. Also, the diagonals of the matrices demonstrate class-wise accuracy, and CP-Mix achieves a fairer class-wise performance by improving accuracy on minority classes. More confusion matrices on different datasets are provided in Appendix~\ref{appendix:experiment:confusion}.

\subsubsection{Class-wise accuracy}
\label{subsubsection:experiments-analysis-classwise-acc}

Figure \ref{figure:analysis}\ref{figure:analysis/comparison-acc-cifar100} and \ref{figure:analysis/comparison-acc-cifar10} compare the class-wise accuracy of the models trained on CIFAR100-LT-200 and CIFAR10-LT-200 datasets. We group adjacent classes into 10 subgroups for visualization of CIFAR100-LT dataset. CP-Mix boosts the performance of minorities by a large margin while sacrificing the performance of majorities little. Balanced softmax, which considers the imbalance and achieves trade-offs between majorities and minorities, demonstrates much lower benefits. As analyzed in Section \ref{subsection:cifar-example}, Mixup improves generalization in majorities by sacrificing minority groups, resulting in a more biased classifier.

\begin{table*}[t!]
  \centering
  \scalebox{1.0}{
  \begin{tabular}{c|c|c|c|c|c|c}
    \toprule
    $l_{ERM}$ & Mixup Reg & Mixing functions & Confusion-pairs & Additional Reg & Finetuning & Top-1 accuracy \\
    \midrule
    CE & $\times$ & $-$ & $-$ & $-$ & $-$ & $34.84$ \\
    BS & $\times$ & $-$ & $-$ & $-$ & $-$ & $38.01$ \\
    \midrule
    BS & $\surd$ & $\times$ & $\times$ & $-$ & $-$ & $36.48$ \\
    BS & $\surd$ & $\surd$ & $\times$ & $-$ & $-$ & $37.21$ \\
    \midrule
    BS & $\surd$ & $\times$ & $\surd$ & $-$ & $-$ & $41.28$ \\
    BS & $\surd$ & $\times$ & $\surd$ & $\surd$ & $-$ & $41.26$ \\
    BS & $\surd$ & $\surd$ & $\surd$ & $-$ & $-$ & $40.26$ \\
    \midrule
    BS & $\surd$ & $\surd$ & $\surd$ & $\surd$ & $\times$ & $42.77$ \\
    BS & $\surd$ & $\surd$ & $\surd$ & $\surd$ & $\surd$ & $43.56$ (CP-Mix) \\
    \bottomrule
  \end{tabular}
  }
  \caption{
  Ablation study to analyze the components of CP-Mix on CIFAR100-LT200 dataset.
  CE and BS denote Cross Entropy and Balanced Softmax, respectively.
  }
  \label{table:ablation:cifar}
\end{table*}

\begin{table*}[t!]
  \centering
  \scalebox{1.0}{
  \begin{tabular}{c|c|c|c|c|c|c}
    \toprule
    $l_{ERM}$ & Mixup Reg & Mixing functions & Confusion-pairs & Additional Reg & Finetuning & Top-1 accuracy \\
    \midrule
    CE & $\times$ & $  -  $  & $  -  $  & $  -  $  & $  -  $  & $45.5$ \\
    BS & $\times$ & $  -  $  & $  -  $  & $  -  $  & $  -  $  & $50.6$ \\
    \midrule
    BS & $\surd$  & $\times$ & $\times$ & $  -  $  & $  -  $  & $46.8$ \\
    BS & $\surd$  & $\surd$  & $\times$ & $  -  $  & $  -  $  & $48.9$ \\
    \midrule
    BS & $\surd$  & $\times$ & $\surd$  & $  -  $  & $  -  $  & $51.8$ \\
    BS & $\surd$  & $\times$ & $\surd$  & $\surd$  & $  -  $  & $51.9$ \\
    BS & $\surd$  & $\surd$  & $\surd$  & $  -  $  & $  -  $  & $52.3$ \\
    \midrule
    BS & $\surd$  & $\surd$  & $\surd$  & $\surd$  & $\times$ & $52.9$ \\
    BS & $\surd$  & $\surd$  & $\surd$  & $\surd$  & $\surd$  & $53.9$ (CP-Mix) \\
    \bottomrule
  \end{tabular}
  }
  \caption{
  Ablation study to analyze the components of CP-Mix on ImageNet-LT dataset.
  CE and BS denote Cross Entropy and Balanced Softmax, respectively.
  }
  \label{table:ablation:imagenetlt}
\end{table*}

\subsubsection{Ablation study}
\label{subsubsection:experiments-analysis-ablation}

We analyze what components of CP-Mix lead to the performance boost in long-tailed image recognition. Table \ref{table:ablation:cifar} and \ref{table:ablation:imagenetlt} present the effects of different components to the performance of CP-Mix for CIFAR100-LT-200 and ImageNet-LT datasets, respectively. First, it is evident that balanced softmax \cite{bs} is a strong baseline that improves cross-entropy by a large margin. Using Mixup regularization and revising mixup formulation to consider imbalance cannot achieve the improvements. By estimating confusion distribution and using confusion-pairs for mixup regularization, we can improve the classifiers by a large margin. Using mixing functions proposed in Section \ref{subsection:cpmixup-mixingfunctions} has limitations but is stabilized with additional regularization, achieving the best performance.

\section{Conclusion}
\label{section:conclusion}

This paper proposes CP-Mix, a novel training framework using mixup to tackle long-tailed recognition.
Inspired by model's tendency to capture class similarities and misclassify minorities to similar majorities, we propose to collect confusion-pairs to augment samples by mixing from these classes.
Our method trains the model to reduce confusion by regularizing to distinguish confusing classes.
We demonstrate the superiority and compatibility of CP-Mix in various long-tailed image classification benchmarks.
Moreover, we analyze CP-Mix and reveal how CP-Mix works and improves classification.
Our work suggests the possibility of Mixup to improve more complex computer vision tasks.


\bibliographystyle{unsrt}  
{
\small
\bibliography{references}
}

\medskip


\appendix

\newpage

\section*{Appendix}

\renewcommand*{\thesection}{\Alph{section}}

\section{Objective function.}
\label{appendix:objective}

CP-Mix augments the objective for ERM training with a regularization objective using confusion pairs. As mentioned in the main paper, we use an additional mixup regularization objective using the same training batch for classification to stabilize the training following RegMix. We use the same mixup hyperparameter $\alpha$ and regularization ratio $\gamma$ to reduce the search space for training hyperparameters. The overall objective for CP-Mix is
\begin{equation}\label{equation:sup-overall}
    {\cal L}(\theta) = {\cal L}_{ERM}(\theta, l_{ERM}) + \gamma_{CP} {\cal L}_{CP} (\theta, l_{CP}) + \gamma_{mix} {\cal L}_{mix} (\theta, l_{mix}),
\end{equation}
where
\begin{equation}\label{equation:sup-erm}
    {\mathcal L}_{ERM}(\theta, l_{ERM}) = {\mathbb E}_{(x, y) \sim {\cal P}}[l_{ERM}(f_\theta (x), y)] \approx {1 \over N} \sum_{(x_i, y_i) \in {\cal D}_s} l_{ERM}(f_\theta (x_i), y_i),
\end{equation}
\begin{equation}\label{equation:sup-cp-objective}
    {\cal L}_{CP}(\theta, l_{CP}) = {\mathbb E}_{(\tilde{x}, \tilde{y}) \sim \tilde{\cal P}_{CP}}[l_{CP}(f_\theta (\tilde{x}), \tilde{y})] \approx {1 \over N''} \sum_{(\tilde{x}_i, \tilde{y}_i) \in {\tilde{\cal D}_{s, CP}}} l_{CP}(f_\theta (\tilde{x}_i), \tilde{y}_i),
\end{equation}
\begin{equation}\label{equation:sup-mixup-objective}
    {\cal L}_{mix}(\theta, l_{mix}) = {\mathbb E}_{(\tilde{x}, \tilde{y}) \sim \tilde{{\cal P}}}[l_{mix}(f_\theta (\tilde{x}), \tilde{y})] \approx {1 \over N'} \sum_{(\tilde{x}_i, \tilde{y}_i) \in {\tilde{\cal D}_s}} l_{mix}(f_\theta (\tilde{x}_i), \tilde{y}_i).
\end{equation}
and Balanced Softmax is used for $l_{ERM}$, and Cross-Entropy is used for $l_{CP}$ and $l_{mix}$. $\tilde{\cal D}_s$ is constructed using the same batch to that of ${\cal D}_s$, and $\tilde{\cal D}_{s, CP}$ is sampled using the class-wise subsets and confusion pairs. CP-Mix utilizes the same mixup strategy for both CP-Mix and regularization objectives.
\section{Implementation details}
\label{appendix:implementation}

\subsection{Toy example}

We use a 2D example to demonstrate how a decision boundary is constructed, and Mixup fails to improve minority generalization.
There are 4 classes where data points are sampled from the Gaussian distributions.
The centers for majorities are $(0, 1)$ and $(1, 0)$, and are $(0, -1)$ and $(-1, 0)$ for minorities.
The standard deviation is $0.4$ for all classes, and $1,000$ and $50$ training points are sampled for majorities and minorities, respectively.
We sampled $1,000$ points for the test set from each class.
The classifier is a fully connected network with one hidden layer containing $100$ nodes (two linear modules), and the ReLU function is used for activation.
Adam optimizer is used, and the classifier is trained for $10$ epochs with a learning rate of $0.1$ and batch size of $100$.

\subsection{Datasets and architecture}
\label{appendix:implementation:dataset}

This subsection presents the details of datasets and network architectures we use to evaluate CP-Mix.
We follow the standard process to construct imbalanced datasets.
The information is summarized in Table \ref{table:sup-datasets}.

\begin{table*}[h!]
  \centering
  \scalebox{1.0}{
  \begin{tabular}{l|cccc}
    \toprule
    Dataset & $\#$ of classes & $\#$ of training samples & Imbalance factor & Architecture \\
    \midrule
    CIFAR-LT          & $100 / 10$   & $50,000$    & $[10, 50, 100, 200]$ & ResNet-32  \\
    ImageNet-LT       & $1,000$      & $115,846$   & $256$                & ResNet-50 \\
    Places-LT         & $365$        & $62,500$    & $996$                & Pre-trained ResNet-152 \\
    iNaturalist 2018  & $8,124$      & $427,513$   & $500$                & ResNet-50 \\
    \bottomrule
  \end{tabular}
  }
  \caption{
  Detailed information of datasets and architecture. Our choices for model architecture are based on previous works.
  }
  \label{table:sup-datasets}
\end{table*}

\begin{table*}[h!]
    \centering
    \scalebox{1.0}{
    \begin{tabular}{l|cccc}
        \toprule
        & CIFAR-LT & ImageNet-LT & Places-LT & iNaturalist 2018 \\
        \midrule
        batch size          & $128$ & $192$ & $128$ & $256$ \\
        max epochs          & $300$ & $180$ & $30$ & $100$ \\
        \midrule
        stage 1 epochs      & $200$ & $120$ & $15$ & $50$ \\
        stage 2 epochs      & $100$ & $60$ & $15$ & $50$ \\
        \midrule
        optimizer           & SGD & SGD & SGD & SGD \\
        learning rate       & $0.1$ & $0.1$ & $0.01$ & $0.1$ \\
        momentum            & $0.9$ & $0.9$ & $0.9$ & $0.9$ \\
        weight decay        & $0.0002$ & $0.0005$ & $0.0005$ & $0.0002$ \\
        \midrule
        scheduler           & warmup multistep & cosine & multistep & warmup multistep \\
        steps               & $[260, 280]$ & $ - $ & $[5, 15]$ & $[60, 80]$ \\
        decay rate          & $0.01$ & $ - $ & $0.1$ & $0.1$ \\
        \midrule
        $t$ & $0.5$ & $0.5$ & $0.5$ & $0.5$ \\
        $\alpha = \alpha_{CP} = \alpha_{mix}$           & $1.5$ & $3.0$ & $1.0$ & $1.0$ \\
        $\gamma = \gamma_{CP} = \gamma_{mix}$       & $2.5$ & $1.0$ & $2.0$ & $1.0$ \\
        \midrule
        fine-tuning epochs  & $30$ & $10$ & $-$ & $10$ \\
        fine-tuning lr      & $0.5$, $0.1$ & $0.05$ & $-$ & $3.0$ \\
        fine-tuning momentum & $0.9$ & $0.9$ & $-$ & $0.9$ \\
        fine-tuning weight decay & $0.0$ & $0.0$ & $-$ & $0.0$ \\
        \bottomrule
    \end{tabular}
    }
    \caption{
    Hyperparameter settings for experimental results.
    We provide more details and settings about RIDE training in the text.
    }
    \label{table:sup-hyperparameters}
\end{table*}

\subsubsection{CIFAR-LT}

CIFAR-LT datasets are the artificially imbalanced version of CIFAR datasets with exponential class imbalance.
Specifically, the number of training examples in class $i$ is reduced according to an exponential function as $n_i = n_0 \mu^i$, and the imbalance factor of the dataset is $\rho = {n_0 \over n_{C-1}} = \mu^{-C+1}$.
We conduct experiments for CIFAR100-LT and CIFAR10-LT datasets with an imbalance factor in $[200, 100, 50, 10]$.
A general augmentation strategy with random crop and random horizontal flip is used after 4 pixels padding for the comparison.
Following previous works, we use ResNet-32 architecture for CIFAR-LT datasets.

\subsubsection{ImageNet-LT}

ImageNet-LT is sampled from ImageNet dataset under the long-tailed Pareto distribution with power value $\alpha = 6$.
It contains $115.8$K examples from $1,000$ classes with the largest and smallest class size $1,260$ and $5$, respectively ($\rho = 256$).
Examples are resized to $256 \times 256$, and random crop, random horizontal flip, and color jitter are used for data augmentation.
We use ResNet-50 architecture for this large-scale dataset following previous works.

\subsubsection{Places-LT}

Places-LT is an imbalanced dataset sampled from a large-scale Places dataset following a strategy similar to ImageNet-LT.
It comprises $184.5$K examples from $365$ classes with the largest and smallest class size $4,980$ and $5$ ($\rho = 996$).
The same pre-processing is used with ImageNet-LT.
Following the previous works, we use ResNet-152 architecture pre-trained using ImageNet.

\subsubsection{iNaturalist 2018}

iNaturalist 2018 dataset is a species classification benchmark with a natural long-tailed class distribution.
It contains $437.5$K examples from $8,142$ classes with the fine-grained hierarchy.
The largest classes have $1,000$ examples, while the smallest has only $2$.
We use ResNet-50 architecture with a normed layer for the classifier following the literature.
We note that a normed layer contributes significantly to the baseline performance.

\subsection{Training details}
\label{appendix:implementation:training}

This subsection details the implementations and hyperparameters for the main results and analysis.
Our experiment results are divided into two parts: base results and RIDE results.
Training details mainly follow previous works for base results, except for the longer training of CIFAR-LT and ImageNet-LT experiments due to additional regularization.
We reduce the batch size if necessary since mixup regularization requires additional memory usage.
For RIDE results, we strictly follow the settings from the original paper for a fair comparison.
We note that RIDE uses a deferred reweighting strategy, and we use six experts for all cases.
Also, we did not fine-tune the classifier for RIDE results.
Basic hyperparameter settings for base results are summarized in Table \ref{table:sup-hyperparameters}, and we detail more specifics and RIDE training settings in the following subsections.

CP-Mix is a two-stage training framework where the first one is similar to the ERM training, and the second stage involves Mixup regularization using confusion-pairs.
The only difference between our first stage and ERM training is that CP-Mix should collect confusion-pairs from the training phase.
Generally, two over three of the whole training time is stage 1, and one over three is for stage 2.

\subsubsection{CIFAR-LT}

We train the model for $300$ epochs with CP-Mix regularization for the last $100$ epochs and a learning rate decays at $260$ and $280$ epochs with $10$ warmup epochs warmup.
For base results, we finetune the models with the backbone frozen for the $30$ epochs.
The learning rate for finetuning is $0.5$ for CIFAR100-LT datasets and $0.1$ for CIFAR10-LT datasets.
For RIDE training, we train the model for $200$ epochs using a normed layer for the classifier, Nesterov momentum for SGD optimizer, and $0.0005$ for weight decay, following the original paper.
We do not finetune the models for RIDE results.

\subsubsection{ImageNet-LT}

We use cosine annealing for base results and train the models for $180$ epochs.
After that, we finetune the classifier for $10$ epochs with the feature extractor frozen.
The learning rate is $0.5$ for finetuning.
Following the original implementation, we train the RIDE models for $100$ epochs with multistep scheduling at $60$ and $80$ epochs with a linear warmup for $5$ epochs.

\subsubsection{Places-LT}

We train the pre-trained model for $30$ epochs with the learning rate decayed at $5$ and $15$ epochs by 0.1, starting from $0.01$.
We do not fine-tune the classifier for Places-LT dataset.

\subsubsection{iNaturalist 2018}

Following the previous works, we use ResNet-50 architecture to evaluate models on iNaturalist 2018 dataset.
We train the models for $100$ epochs with the learning rate $0.1$, decayed at $60$ and $80$ epochs by 0.1 with $10$ linear warmup epochs.
We also note that the value of weight decay (L2 regularization) matters a lot in the overall performance of this dataset, and we use a normed layer as a classifier following the literature.
We finetune the classifier for $10$ epochs with the learning rate annealed by cosine scheduling.
The learning rate for fine-tuning is $3.0$.

\section{Additional experimental results}
\label{appendix:experiment}

This section provides additional results of our implementation for the main results and subgroup performance.
Then we also present additional results for more analysis.

\subsection{Main results}
\label{appendix:experiment:results}

\subsubsection{ImageNet-LT}

We report three sub-groups performance (many, medium, and few) for ImageNet-LT dataset.
Results are summarized in Table \ref{table:sup-main-imagenet-res50}.
While the performance of CP-Mix for Many sub-group is dropped by about $2 \%$ compared to balanced softmax, it significantly improves in Medium and Few sub-groups.

\begin{table*}[h!]
  \centering
  \scalebox{1.0}{
  \begin{tabular}{c|ccc|c}
    \toprule
    Method & Many & Medium & Few & All \\
    \midrule
    Cross Entropy$^\dagger$           & $65.9$ & $37.6$ & $9.1$ & $44.6$ \\
    LDAM-DRW$^\dagger$                & $62.1$ & $46.4$ & $30.1$ & $50.3$ \\
    Balanced Softmax$^\dagger$        & $61.6$ & $48.1$ & $30.3$ & $50.9$ \\
    MiSLAS & $61.7$ & $51.3$ & $35.8$ & $52.7$ \\
    \midrule
    CP-Mix$^\dagger$                  & $59.3$ & $53.5$ & $39.7$ & $53.9$ \\
    \bottomrule
  \end{tabular}
  }
  \caption{
  ImageNet-LT.
  $\dagger$ denotes results reproduced by us, and others are from the original papers.
  }
  \label{table:sup-main-imagenet-res50}
\end{table*}

\subsubsection{Places-LT}

Table \ref{table:sup-main-placeslt} summarizes the results for Places-LT dataset.
CP-Mix achieves significant improvement in overall performance compared to other baselines.

\begin{table*}[h!]
  \centering
  \scalebox{1.0}{
  \begin{tabular}{c|ccc|c}
    \toprule
    Method & Many & Medium & Few & All \\
    \midrule
    Cross Entropy$^\dagger$           & $46.7$ & $28.5$ & $15.5$ & $32.3$ \\ 
    Balanced Softmax$^\dagger$        & $43.2$ & $40.2$ & $31.1$ & $39.3$ \\ 
    MiSLAS & $39.6$ & $43.3$ & $36.1$ & $40.4$ \\
    \midrule
    CP-Mix$^\dagger$                  & $45.4$ & $42.7$ & $36.1$ & $42.3$ \\ 
    \bottomrule
  \end{tabular}
  }
  \caption{
  Places-LT.
  $\dagger$ denotes results reproduced by us, and others are from the original papers.
  }
  \label{table:sup-main-placeslt}
\end{table*}

\subsubsection{iNaturalist 2018}

We report the result for iNaturalist 2018 dataset in Table \ref{table:sup-main-inat18}.
Although the performance drop of CP-Mix for Many sub-group seems significant, we note that the number of classes in Many sub-group is 842 among 8142 classes.
Accordingly, CP-Mix achieves improvement for many classes at the cost of a decrease in a small number of classes.
We note that weight decay affects the results significantly, with about $5 \%$ drop with $0.0005$.

\begin{table*}[h!]
  \centering
  \scalebox{1.0}{
  \begin{tabular}{c|ccc|c}
    \toprule
    Method & Many & Medium & Few & All \\
    \midrule
    Cross Entropy$^\dagger$           & $76.4$ & $66.4$ & $59.1$ & $64.5$ \\ 
    LDAM-DRW$^\dagger$                & $68.3$ & $67.3$ & $66.4$ & $66.9$ \\
    Balanced Softmax$^\dagger$        & $68.8$ & $70.3$ & $69.7$ & $69.9$ \\
    \midrule
    CP-Mix$^\dagger$                  & $57.9$ & $72.9$ & $75.3$ & $72.3$ \\ 
    \bottomrule
  \end{tabular}
  }
  \caption{
  iNaturalist 2018.
  $\dagger$ denotes results reproduced by us, and others are from the original papers.
  }
  \label{table:sup-main-inat18}
\end{table*}

\subsection{Confusion matrix}
\label{appendix:experiment:confusion}

\begin{figure*}[h!]
    \centering
    \begin{subfigure}[b]{0.32\textwidth}
        \centering
        \includegraphics[width=1.0\textwidth]{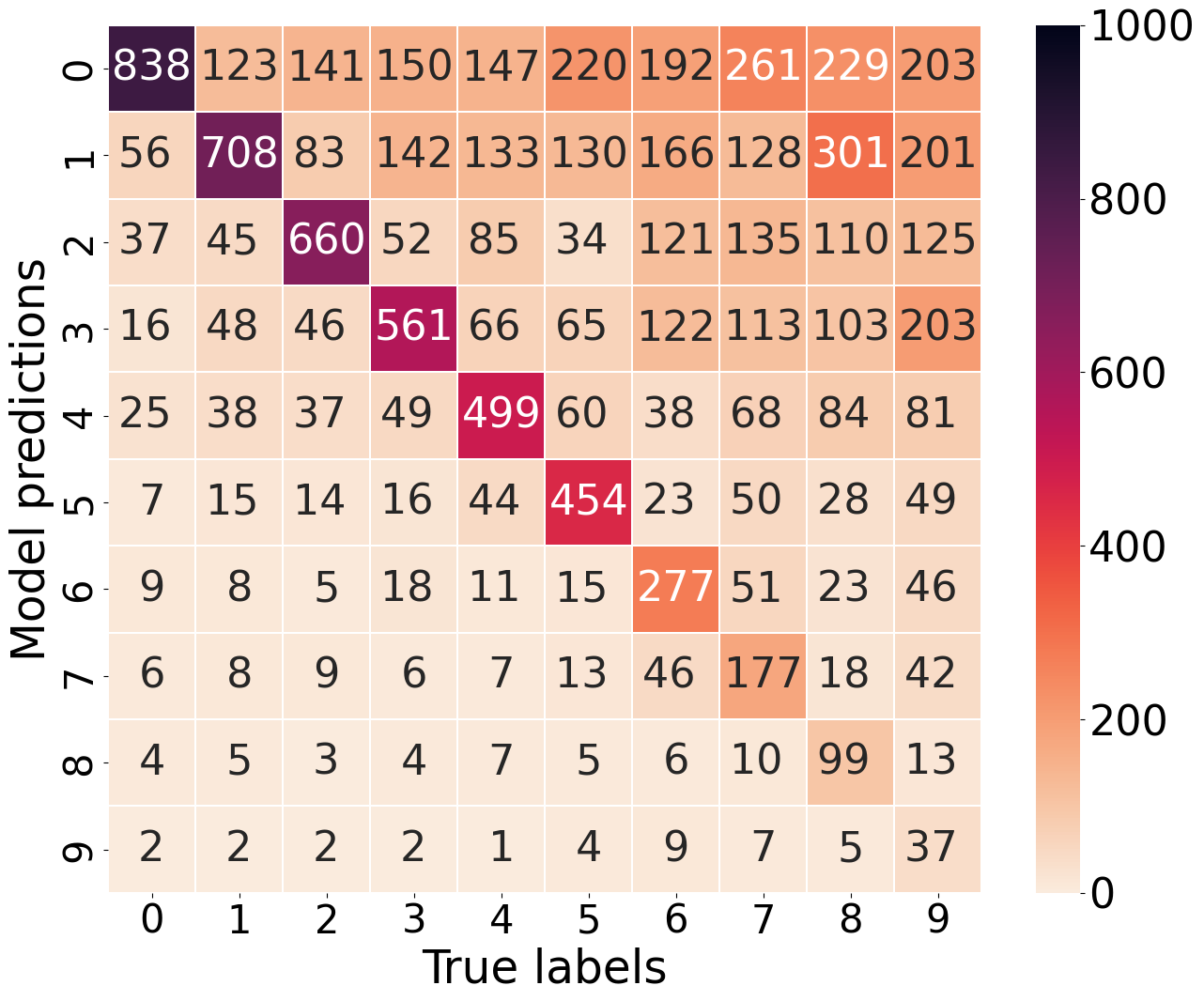}
        \subcaption{ERM, CIFAR100-LT-100 dataset.}
        \label{figure:appendix_confusion_matrix_erm_cifar100_lt_100}
    \end{subfigure}
    \hfill
    \begin{subfigure}[b]{0.32\textwidth}
        \centering
        \includegraphics[width=1.0\textwidth]{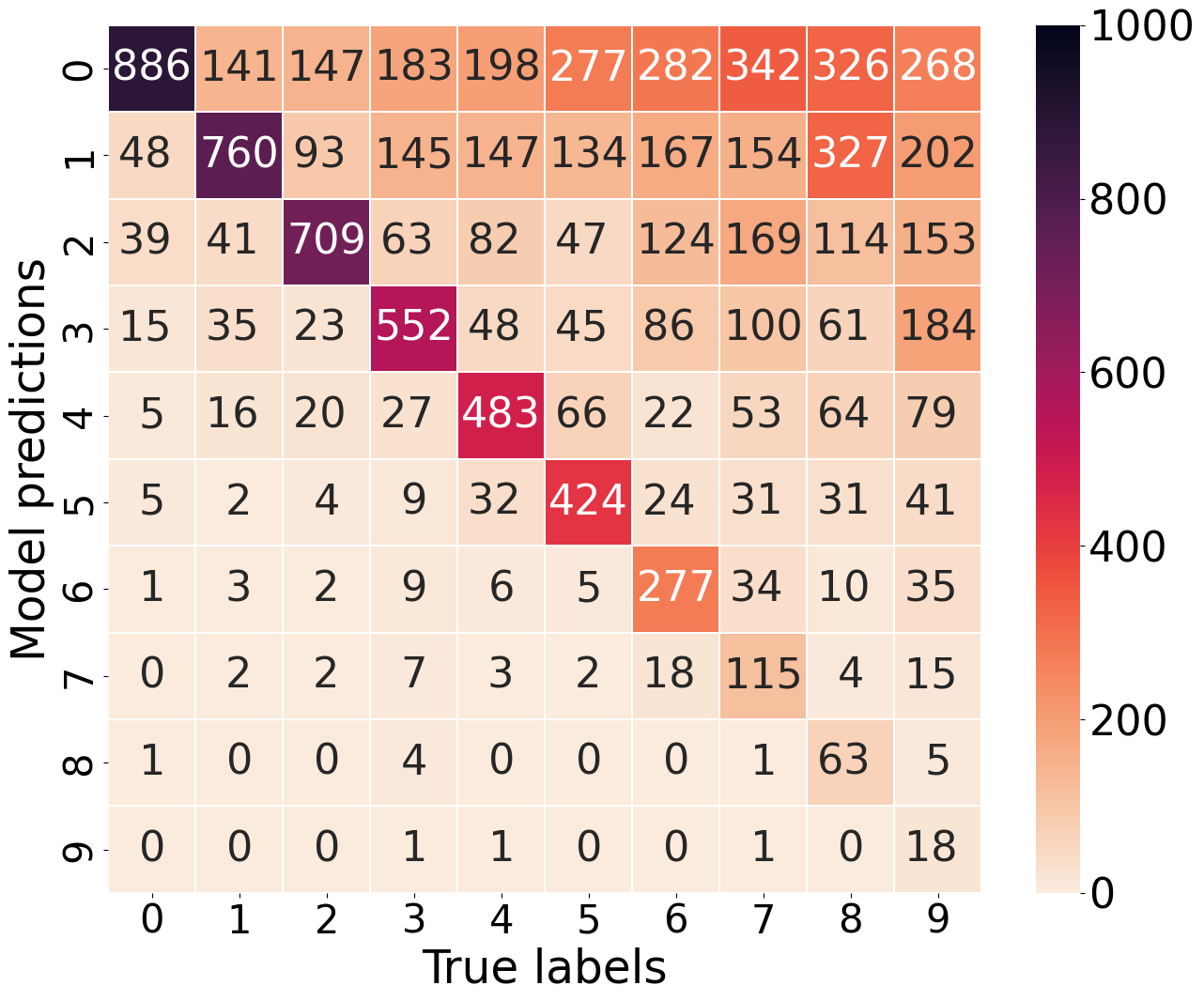}
        \subcaption{Mixup, CIFAR100-LT-100 dataset.}
        \label{figure:appendix_confusion_matrix_mixup_cifar100_lt_100}
    \end{subfigure}
    \hfill
    \begin{subfigure}[b]{0.32\textwidth}
        \centering
        \includegraphics[width=1.0\textwidth]{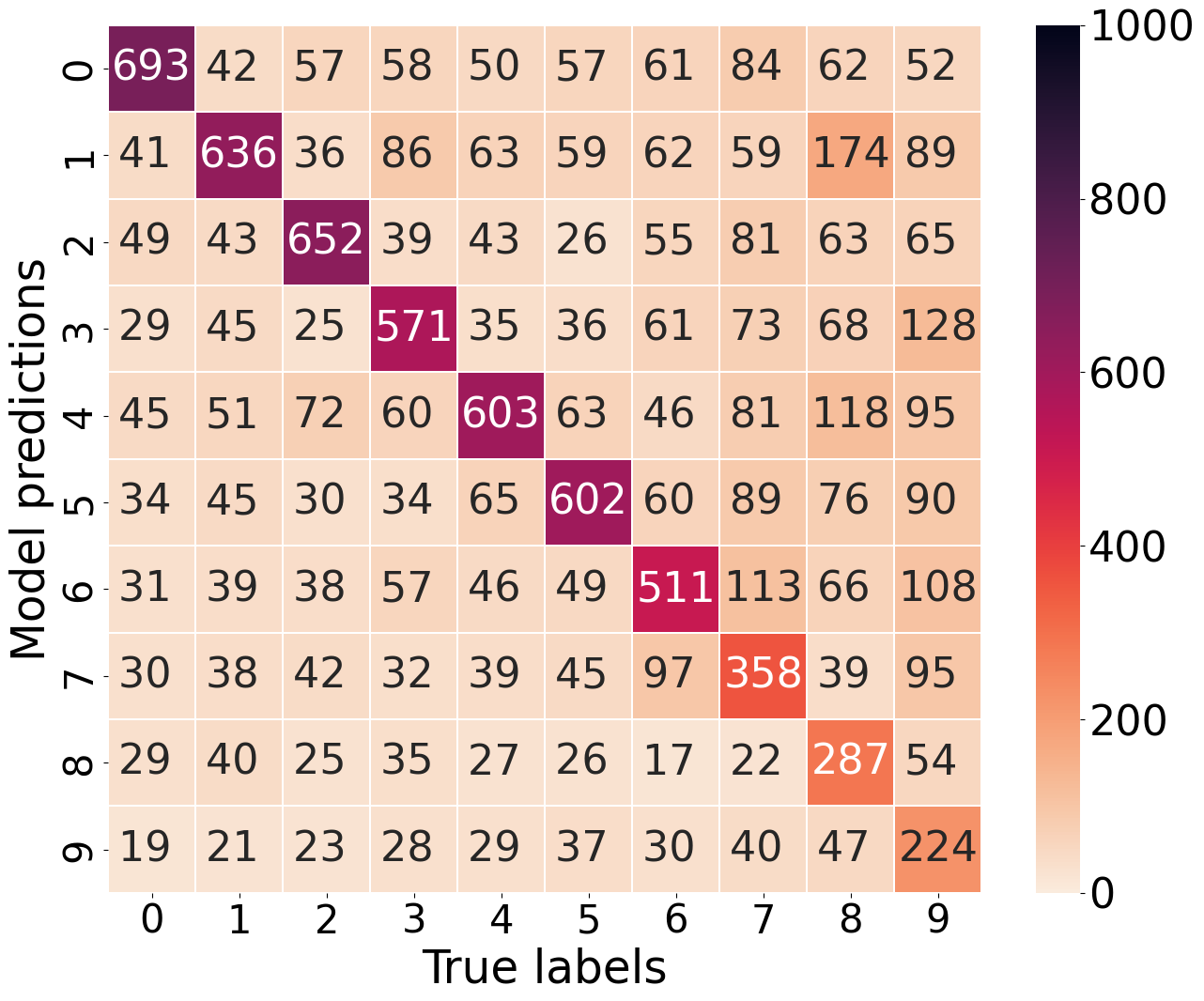}
        \subcaption{CP-Mix, CIFAR100-LT-100 dataset.}
        \label{figure:appendix_confusion_matrix_cpmix_cifar100_lt_100}
    \end{subfigure}
    \hfill

    \begin{subfigure}[b]{0.32\textwidth}
        \centering
        \includegraphics[width=1.0\textwidth]{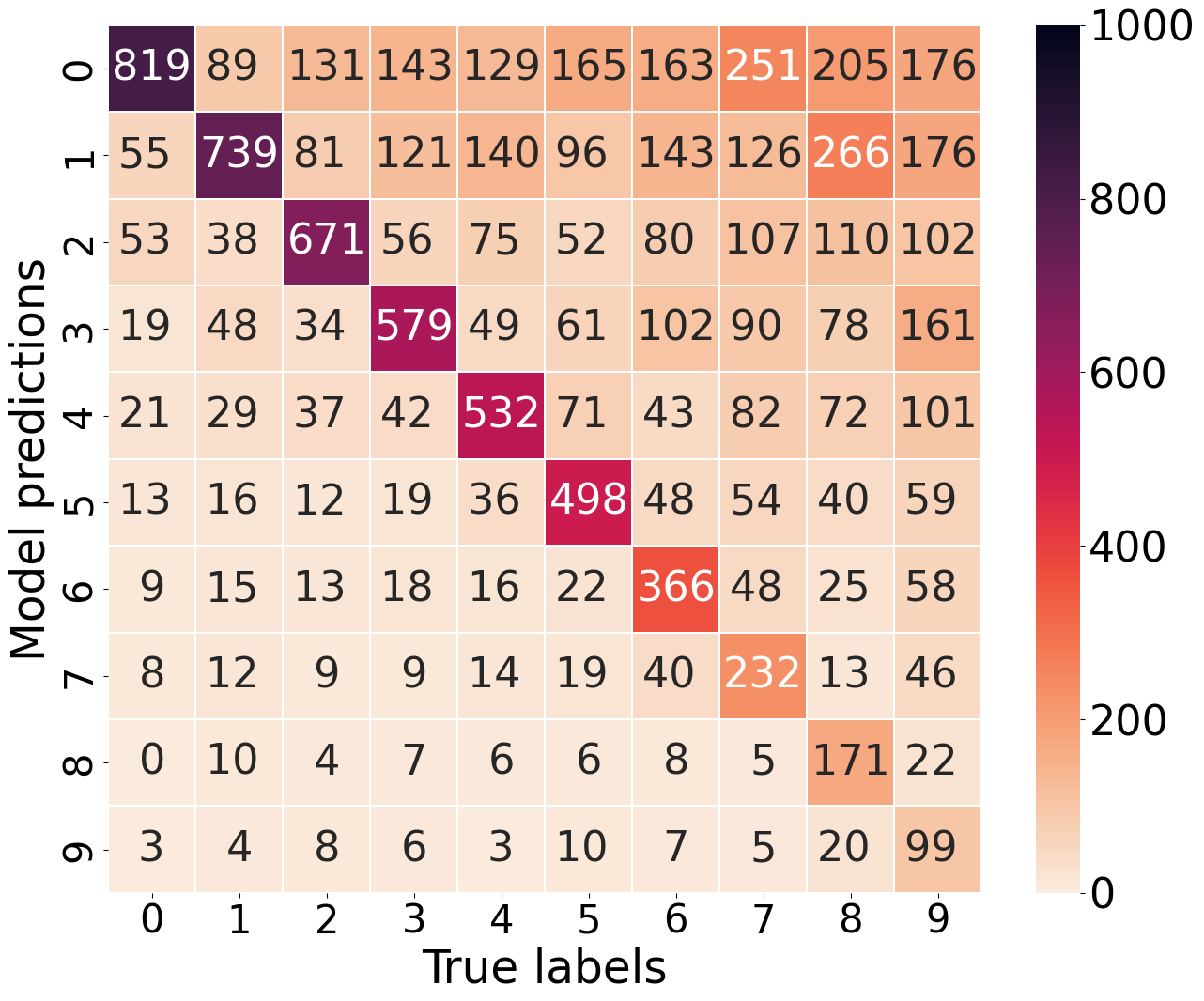}
        \subcaption{ERM, CIFAR100-LT-50 dataset.}
        \label{figure:appendix_confusion_matrix_erm_cifar100_lt_50}
    \end{subfigure}
    \hfill
    \begin{subfigure}[b]{0.32\textwidth}
        \centering
        \includegraphics[width=1.0\textwidth]{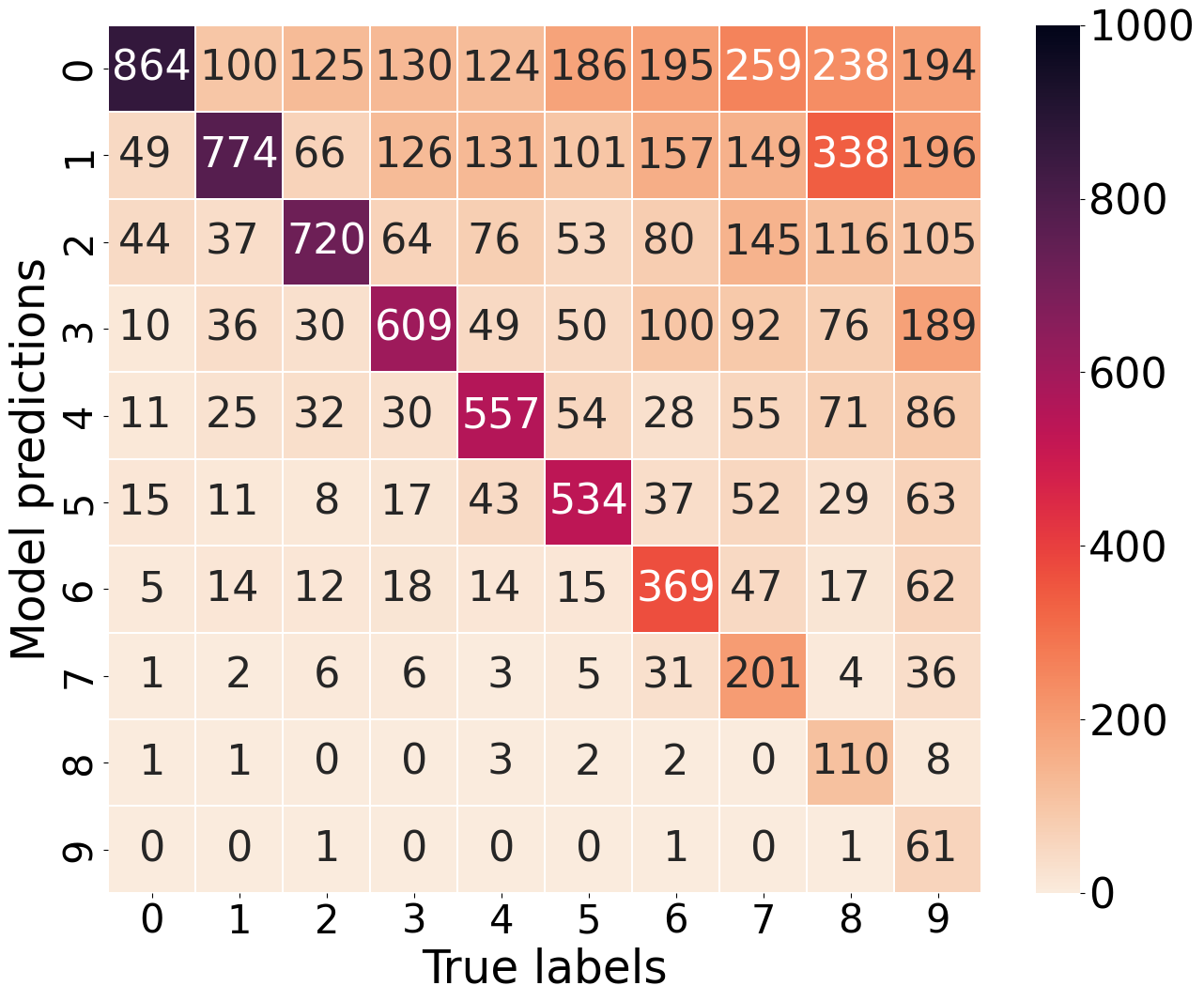}
        \subcaption{Mixup, CIFAR100-LT-50 dataset.}
        \label{figure:appendix_confusion_matrix_mixup_cifar100_lt_50}
    \end{subfigure}
    \hfill
    \begin{subfigure}[b]{0.32\textwidth}
        \centering
        \includegraphics[width=1.0\textwidth]{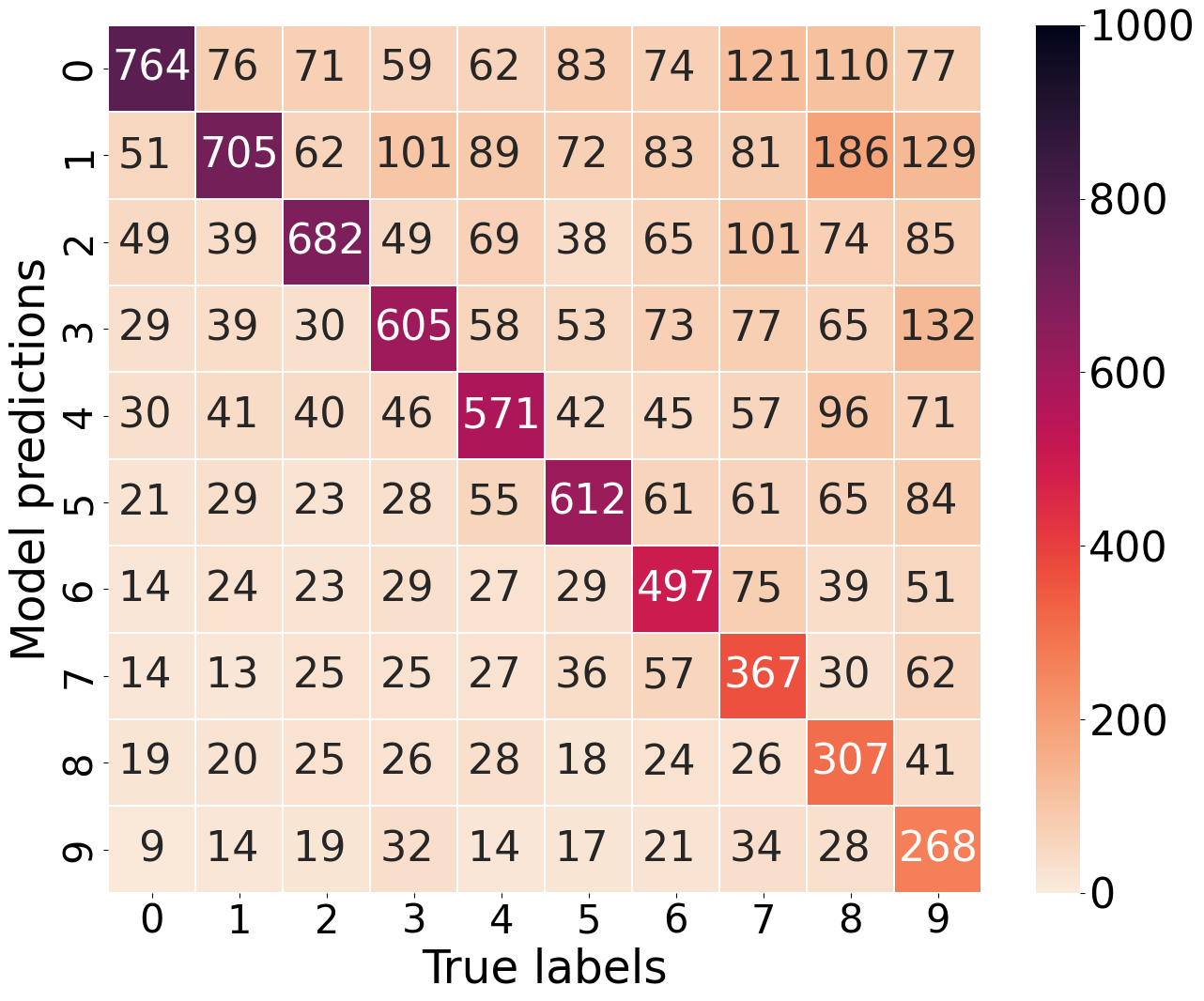}
        \subcaption{CP-Mix, CIFAR100-LT-50 dataset.}
        \label{figure:appendix_confusion_matrix_cpmix_cifar100_lt_50}
    \end{subfigure}
    \hfill

    \begin{subfigure}[b]{0.32\textwidth}
        \centering
        \includegraphics[width=1.0\textwidth]{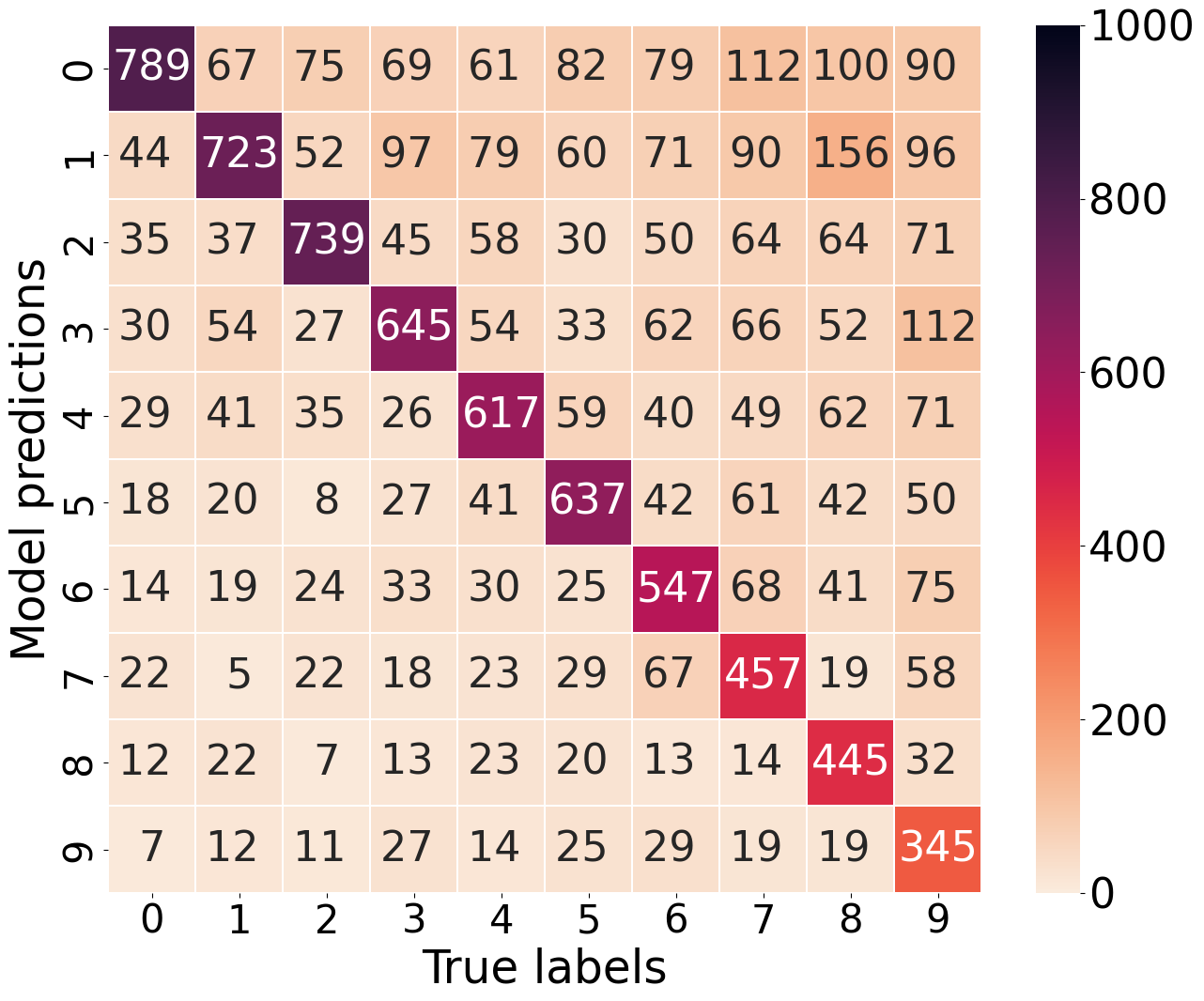}
        \subcaption{ERM, CIFAR100-LT-10 dataset.}
        \label{figure:appendix_confusion_matrix_erm_cifar100_lt_10}
    \end{subfigure}
    \hfill
    \begin{subfigure}[b]{0.32\textwidth}
        \centering
        \includegraphics[width=1.0\textwidth]{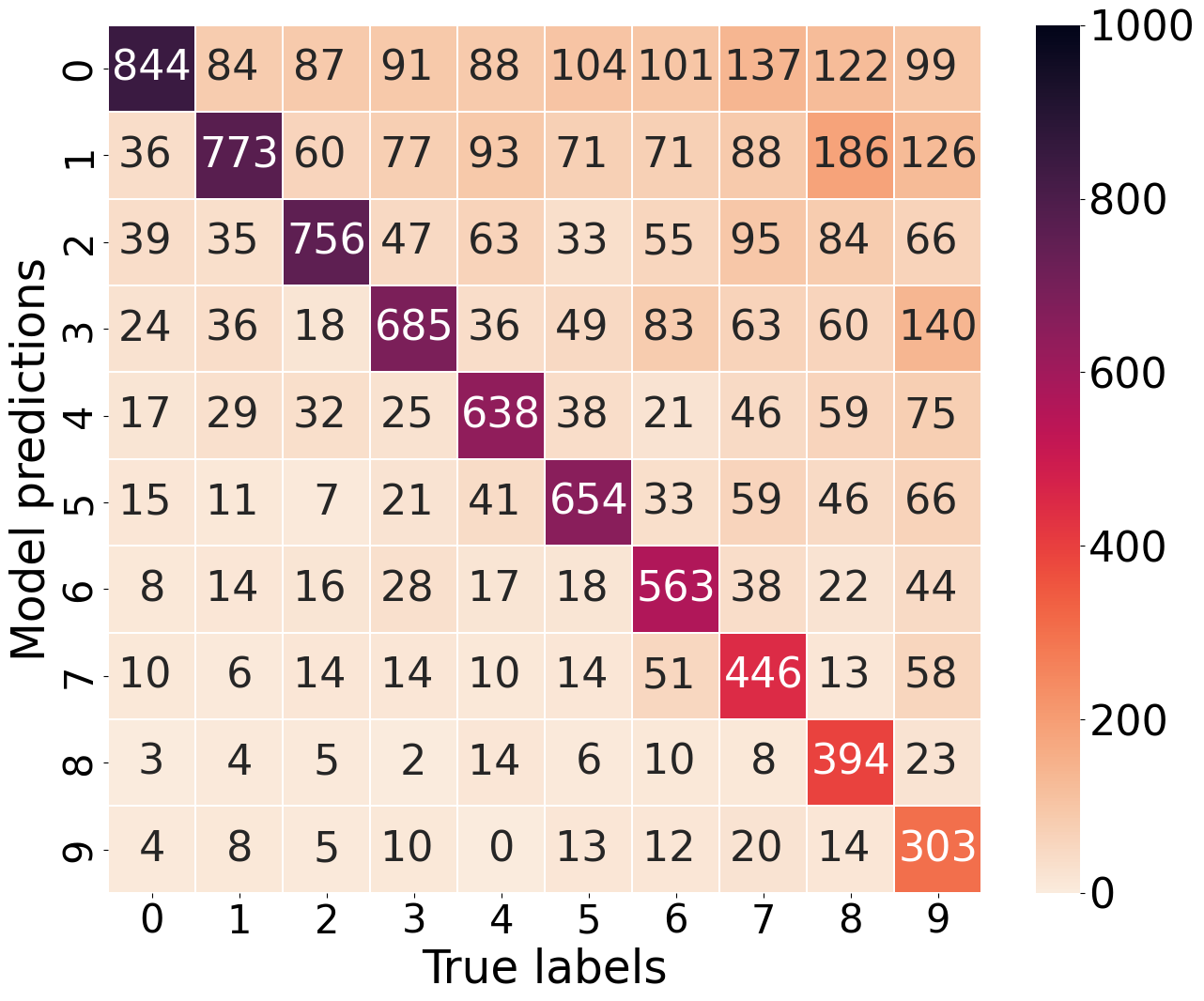}
        \subcaption{Mixup, CIFAR100-LT-10 dataset.}
        \label{figure:appendix_confusion_matrix_mixup_cifar100_lt_10}
    \end{subfigure}
    \hfill
    \begin{subfigure}[b]{0.32\textwidth}
        \centering
        \includegraphics[width=1.0\textwidth]{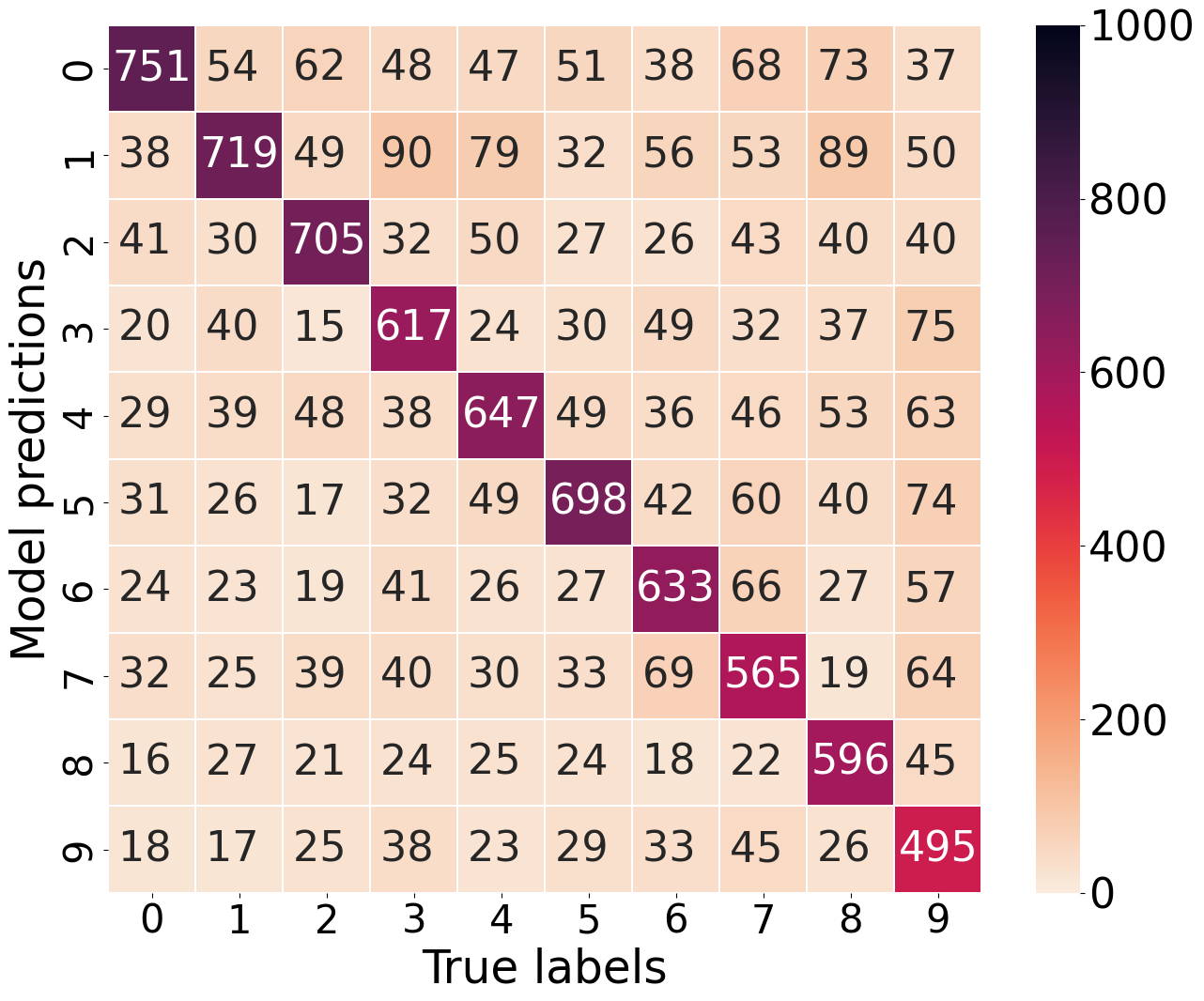}
        \subcaption{CP-Mix, CIFAR100-LT-10 dataset.}
        \label{figure:appendix_confusion_matrix_cpmix_cifar100_lt_10}
    \end{subfigure}
    \hfill
    \caption{
    Confusion matrices of the ERM, Mixup and CP-Mix classifiers trained on CIFAR100-LT datasets.
    }
    \label{figure:appendix_confusion_matrices_cifar100}
\end{figure*}

\begin{figure*}[h!]
    \centering
    \begin{subfigure}[b]{0.32\textwidth}
        \centering
        \includegraphics[width=1.0\textwidth]{figs/cifars/confusion_matrix_cifar10_lt_200_CE.png}
        \subcaption{ERM, cifar10-LT-200 dataset.}
        \label{figure:appendix_confusion_matrix_erm_cifar10_lt_200}
    \end{subfigure}
    \hfill
    \begin{subfigure}[b]{0.32\textwidth}
        \centering
        \includegraphics[width=1.0\textwidth]{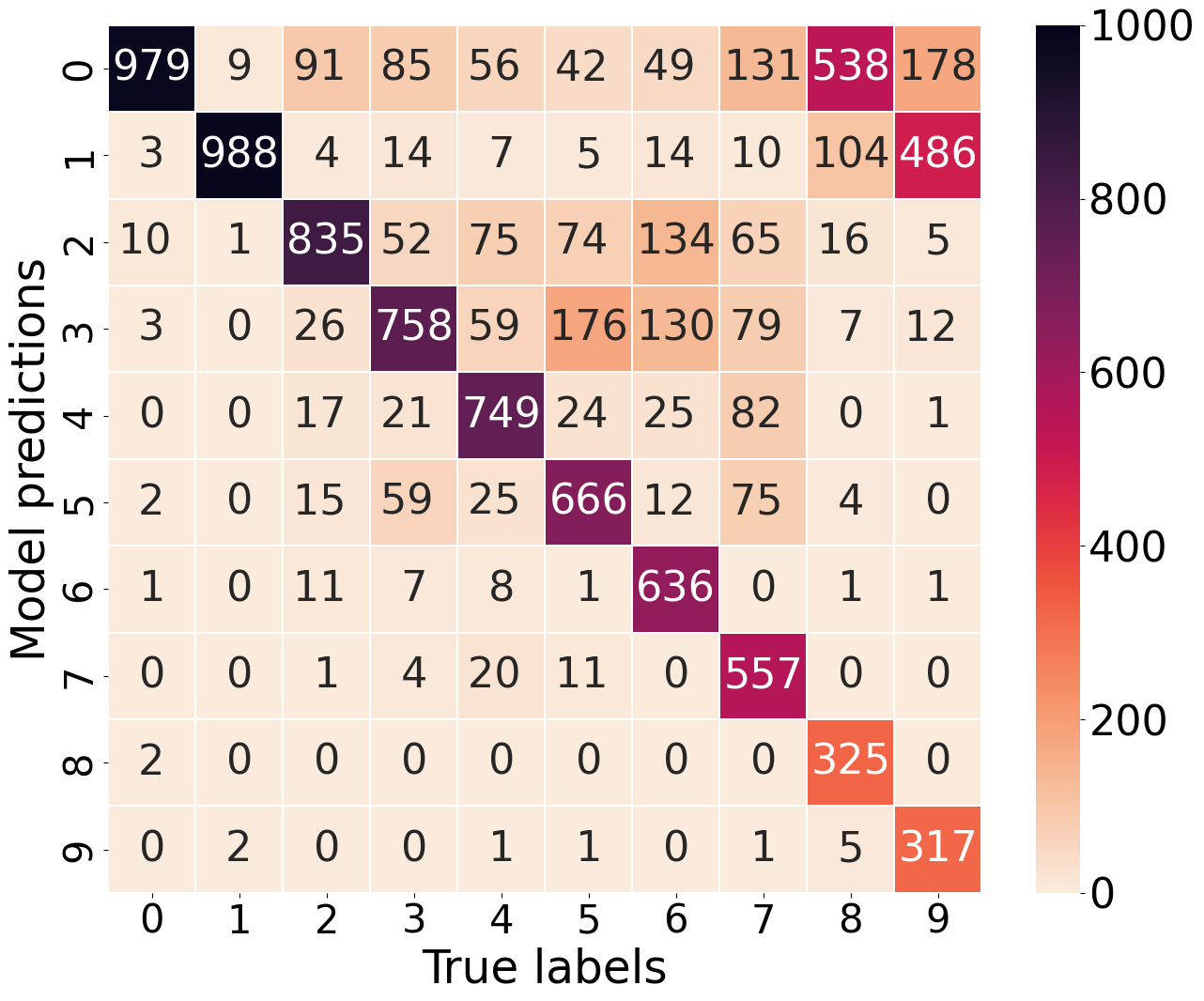}
        \subcaption{Mixup, cifar10-LT-200 dataset.}
        \label{figure:appendix_confusion_matrix_mixup_cifar10_lt_200}
    \end{subfigure}
    \hfill
    \begin{subfigure}[b]{0.32\textwidth}
        \centering
        \includegraphics[width=1.0\textwidth]{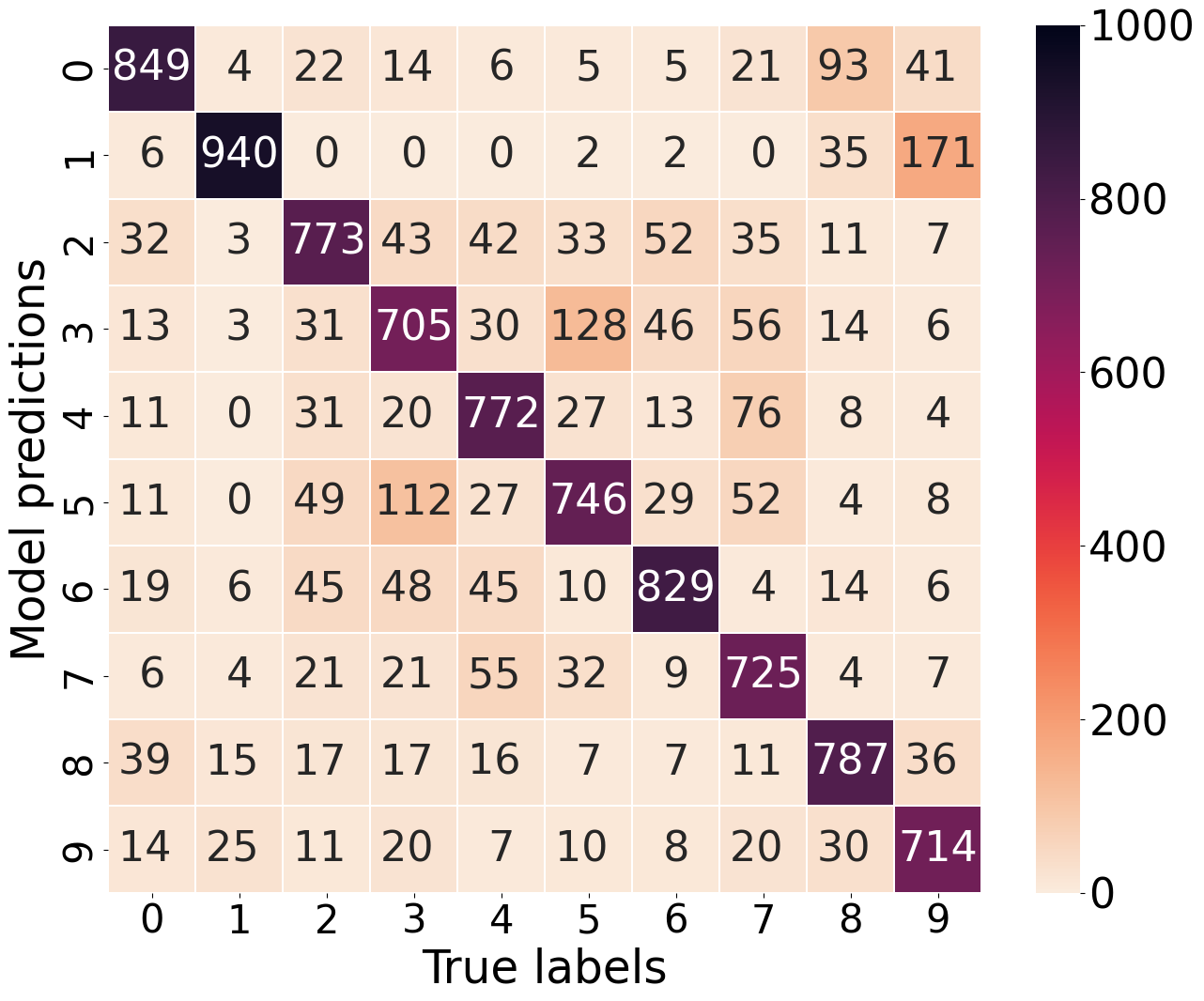}
        \subcaption{CP-Mix, cifar10-LT-200 dataset.}
        \label{figure:appendix_confusion_matrix_cpmix_cifar10_lt_200}
    \end{subfigure}
    \hfill
    
    \begin{subfigure}[b]{0.32\textwidth}
        \centering
        \includegraphics[width=1.0\textwidth]{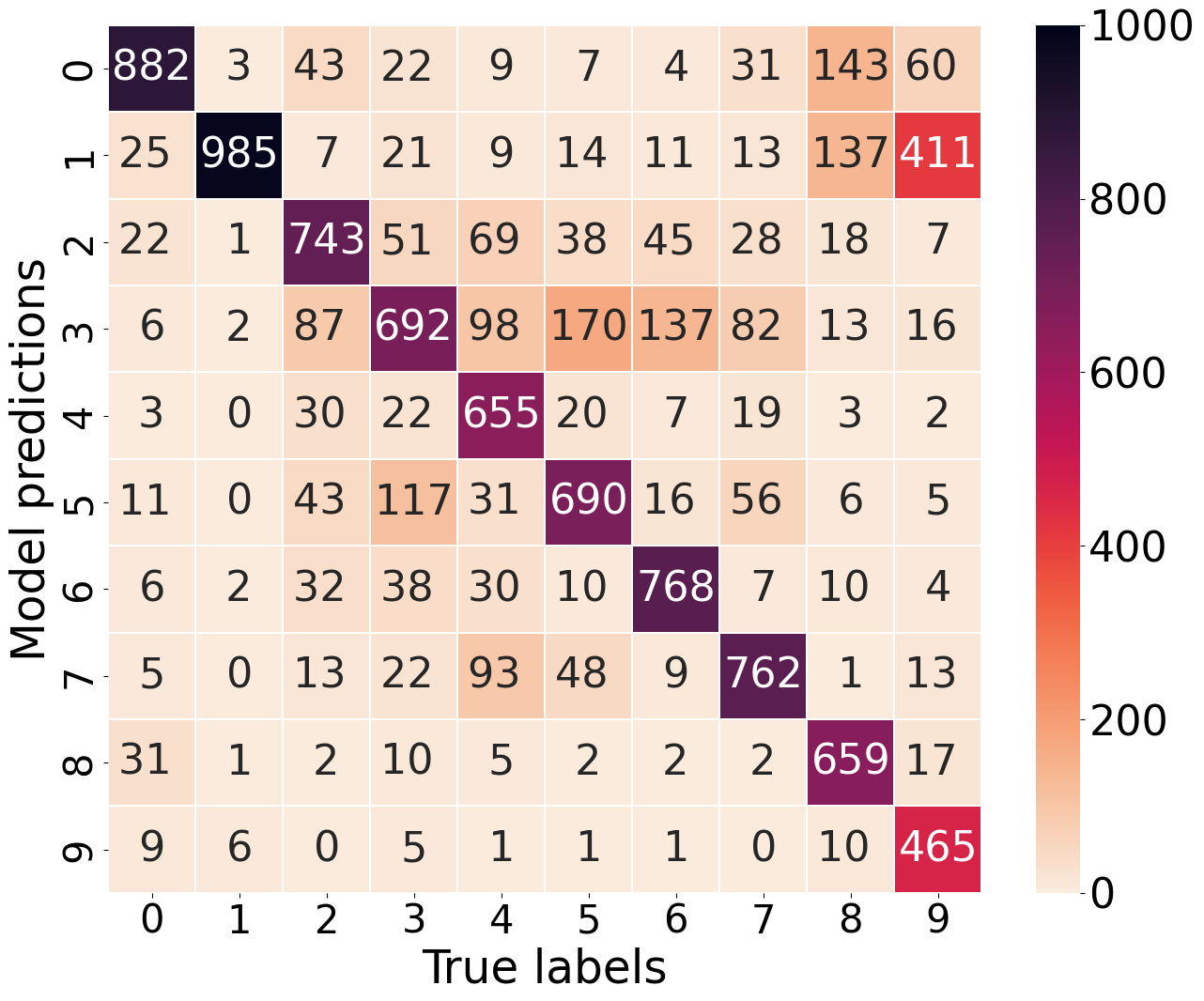}
        \subcaption{ERM, cifar10-LT-100 dataset.}
        \label{figure:appendix_confusion_matrix_erm_cifar10_lt_100}
    \end{subfigure}
    \hfill
    \begin{subfigure}[b]{0.32\textwidth}
        \centering
        \includegraphics[width=1.0\textwidth]{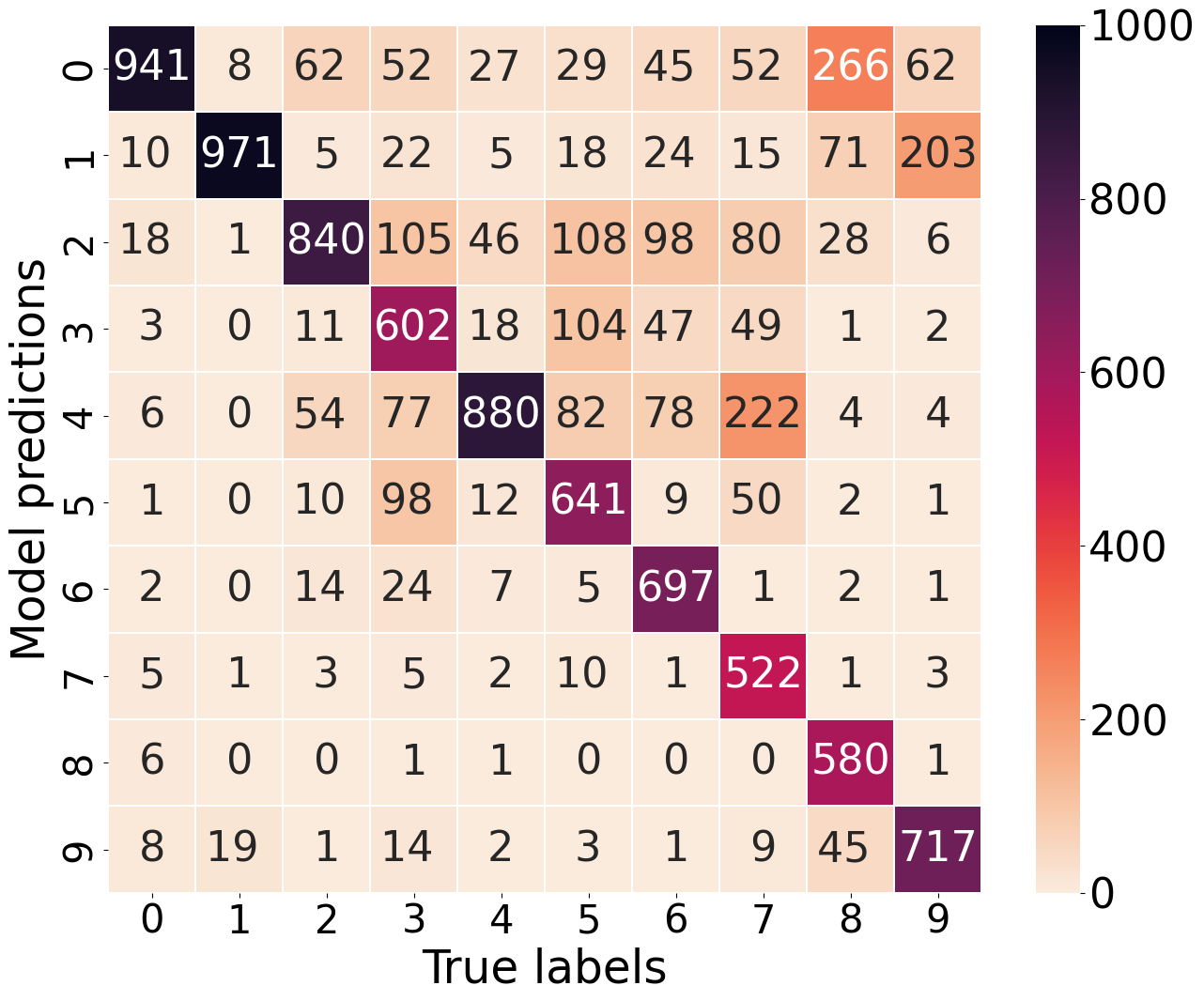}
        \subcaption{Mixup, cifar10-LT-100 dataset.}
        \label{figure:appendix_confusion_matrix_mixup_cifar10_lt_100}
    \end{subfigure}
    \hfill
    \begin{subfigure}[b]{0.32\textwidth}
        \centering
        \includegraphics[width=1.0\textwidth]{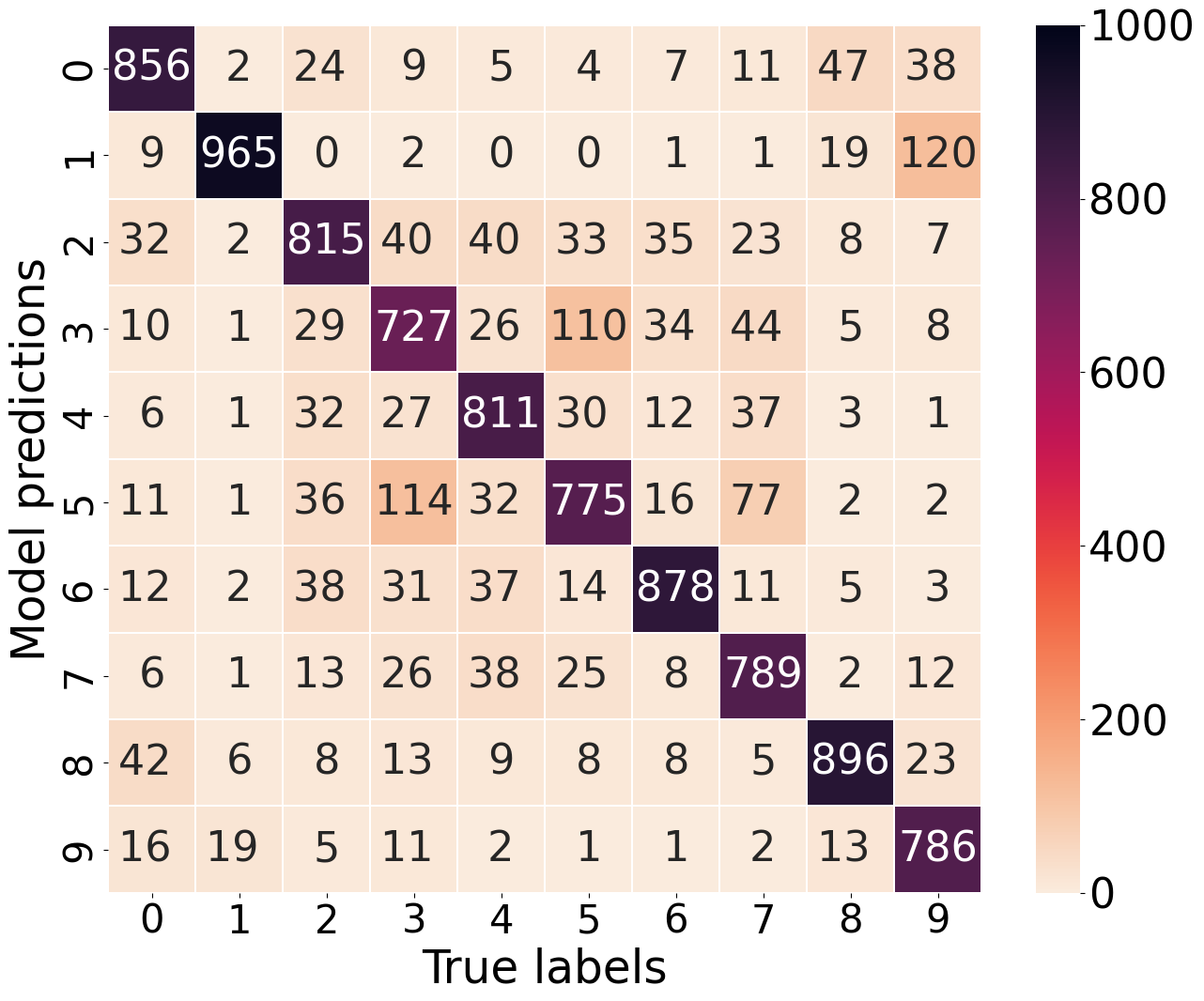}
        \subcaption{CP-Mix, cifar10-LT-100 dataset.}
        \label{figure:appendix_confusion_matrix_cpmix_cifar10_lt_100}
    \end{subfigure}
    \hfill

    \begin{subfigure}[b]{0.32\textwidth}
        \centering
        \includegraphics[width=1.0\textwidth]{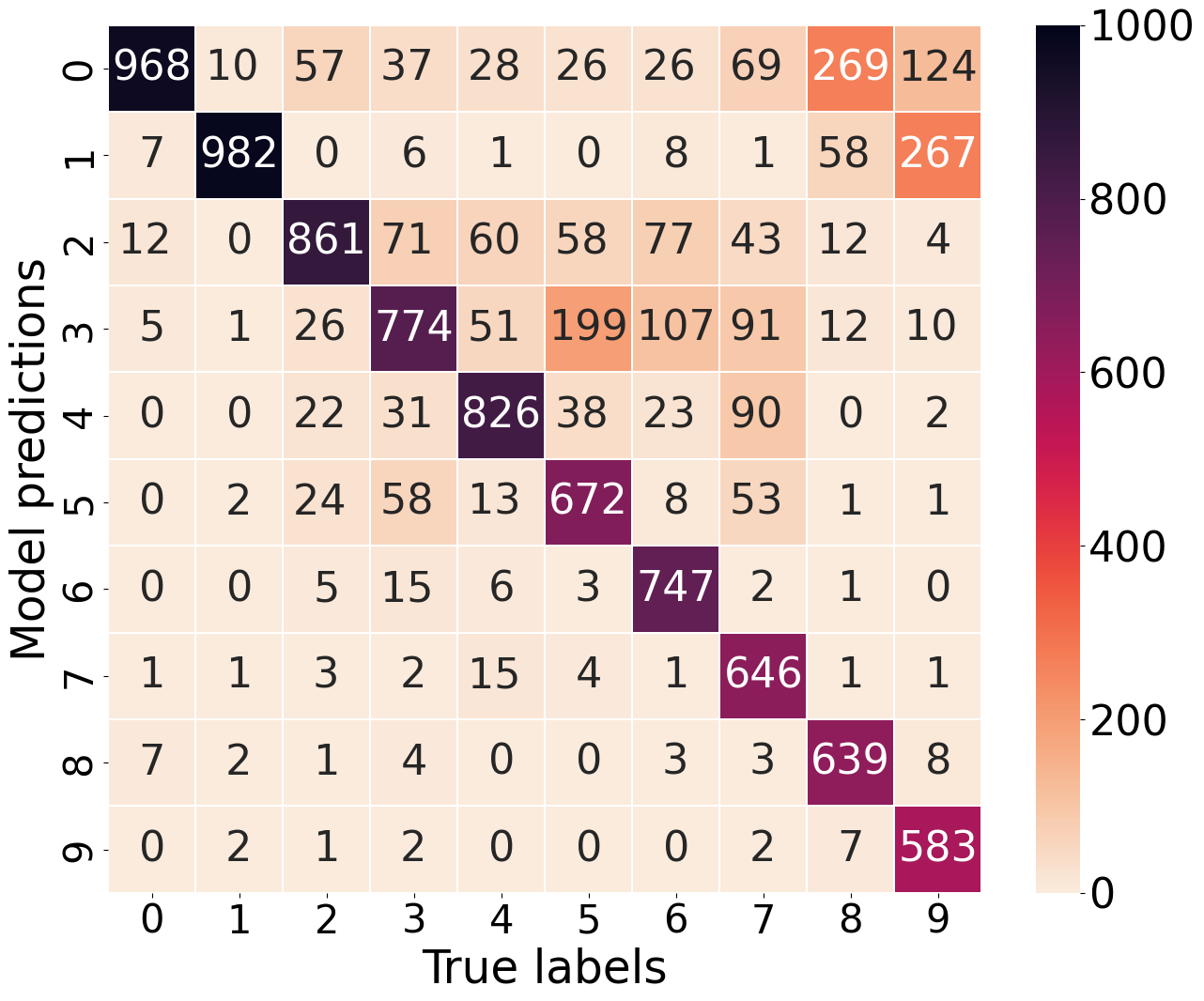}
        \subcaption{ERM, cifar10-LT-50 dataset.}
        \label{figure:appendix_confusion_matrix_erm_cifar10_lt_50}
    \end{subfigure}
    \hfill
    \begin{subfigure}[b]{0.32\textwidth}
        \centering
        \includegraphics[width=1.0\textwidth]{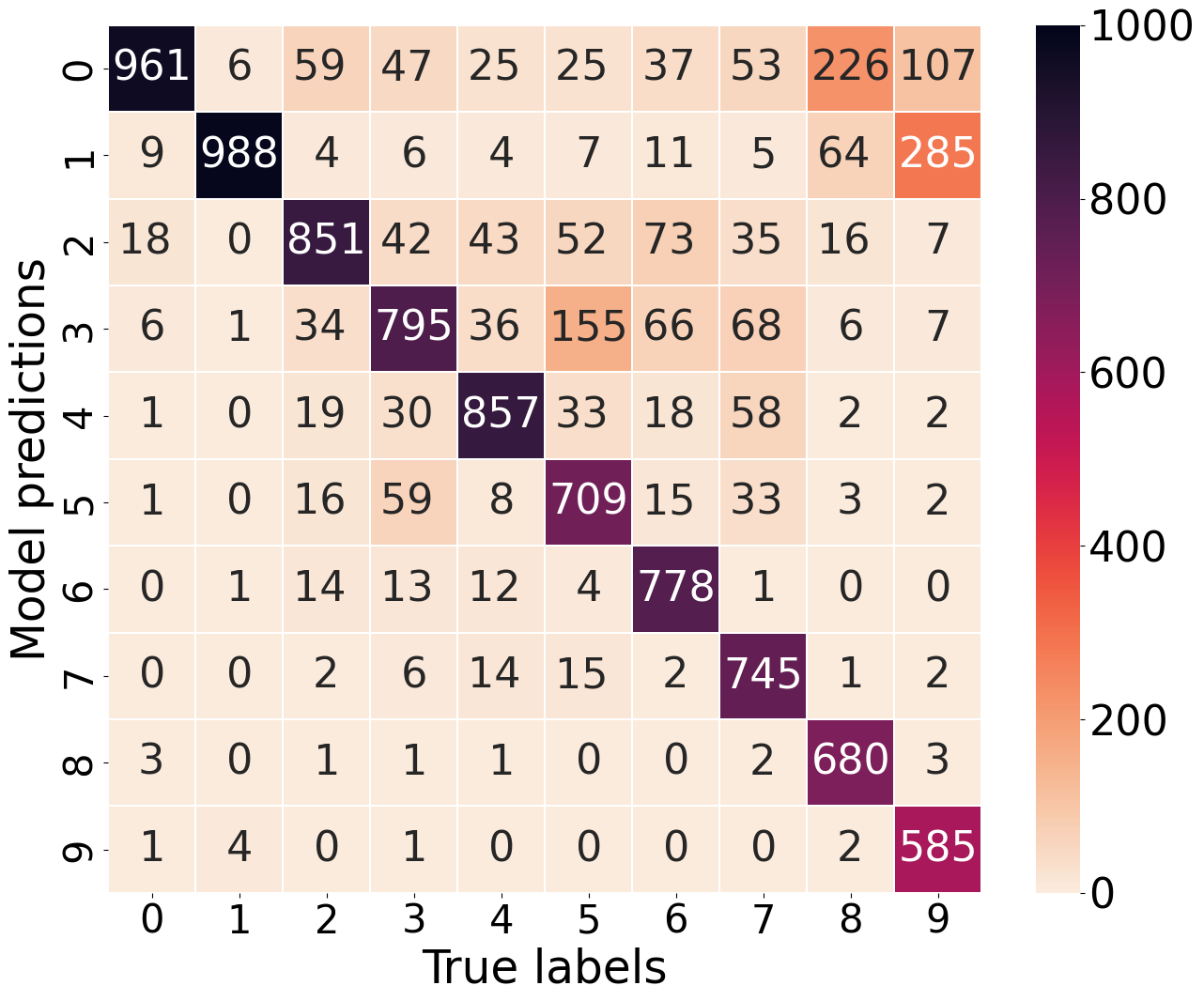}
        \subcaption{Mixup, cifar10-LT-50 dataset.}
        \label{figure:appendix_confusion_matrix_mixup_cifar10_lt_50}
    \end{subfigure}
    \hfill
    \begin{subfigure}[b]{0.32\textwidth}
        \centering
        \includegraphics[width=1.0\textwidth]{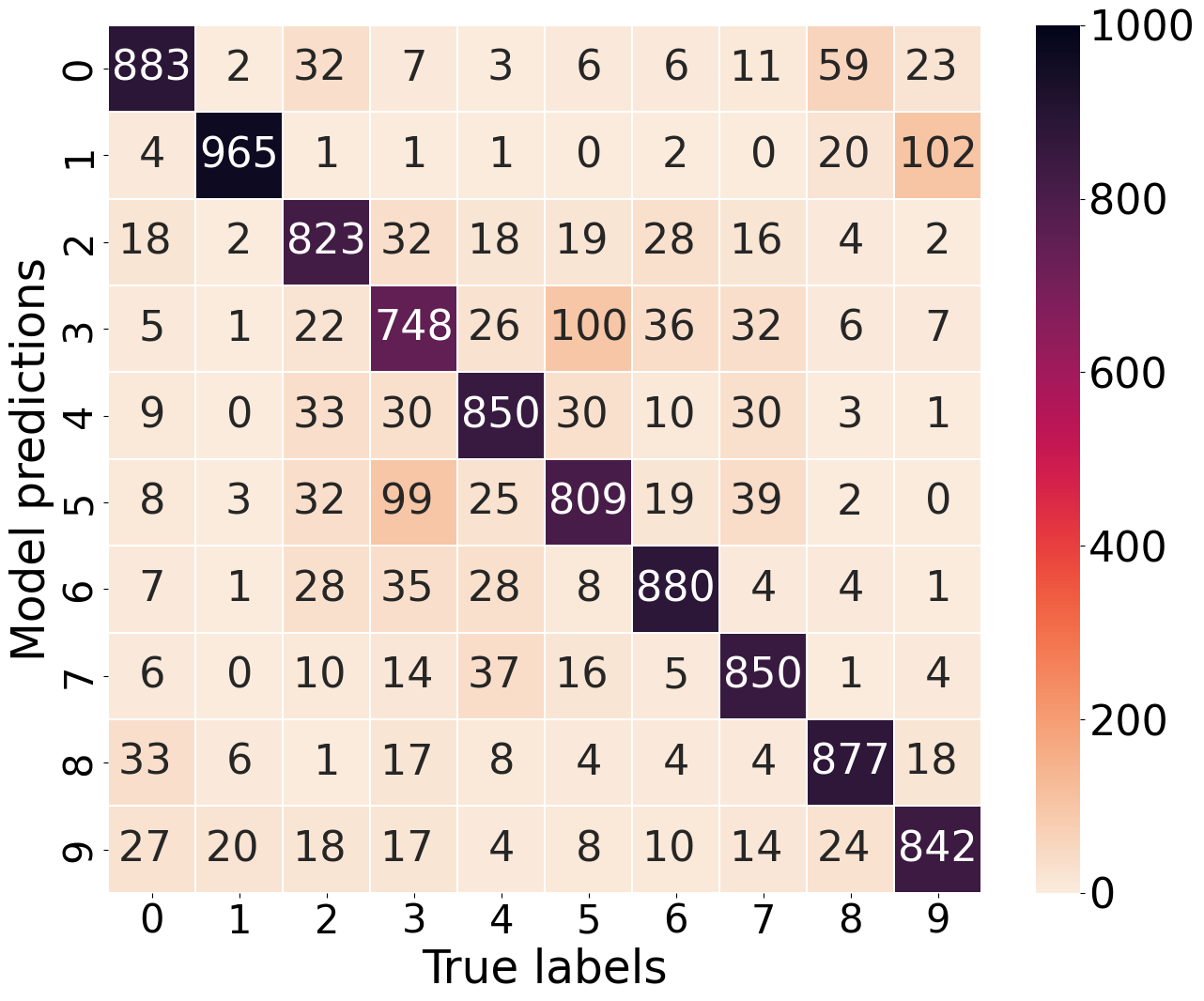}
        \subcaption{CP-Mix, cifar10-LT-50 dataset.}
        \label{figure:appendix_confusion_matrix_cpmix_cifar10_lt_50}
    \end{subfigure}
    \hfill

    \begin{subfigure}[b]{0.32\textwidth}
        \centering
        \includegraphics[width=1.0\textwidth]{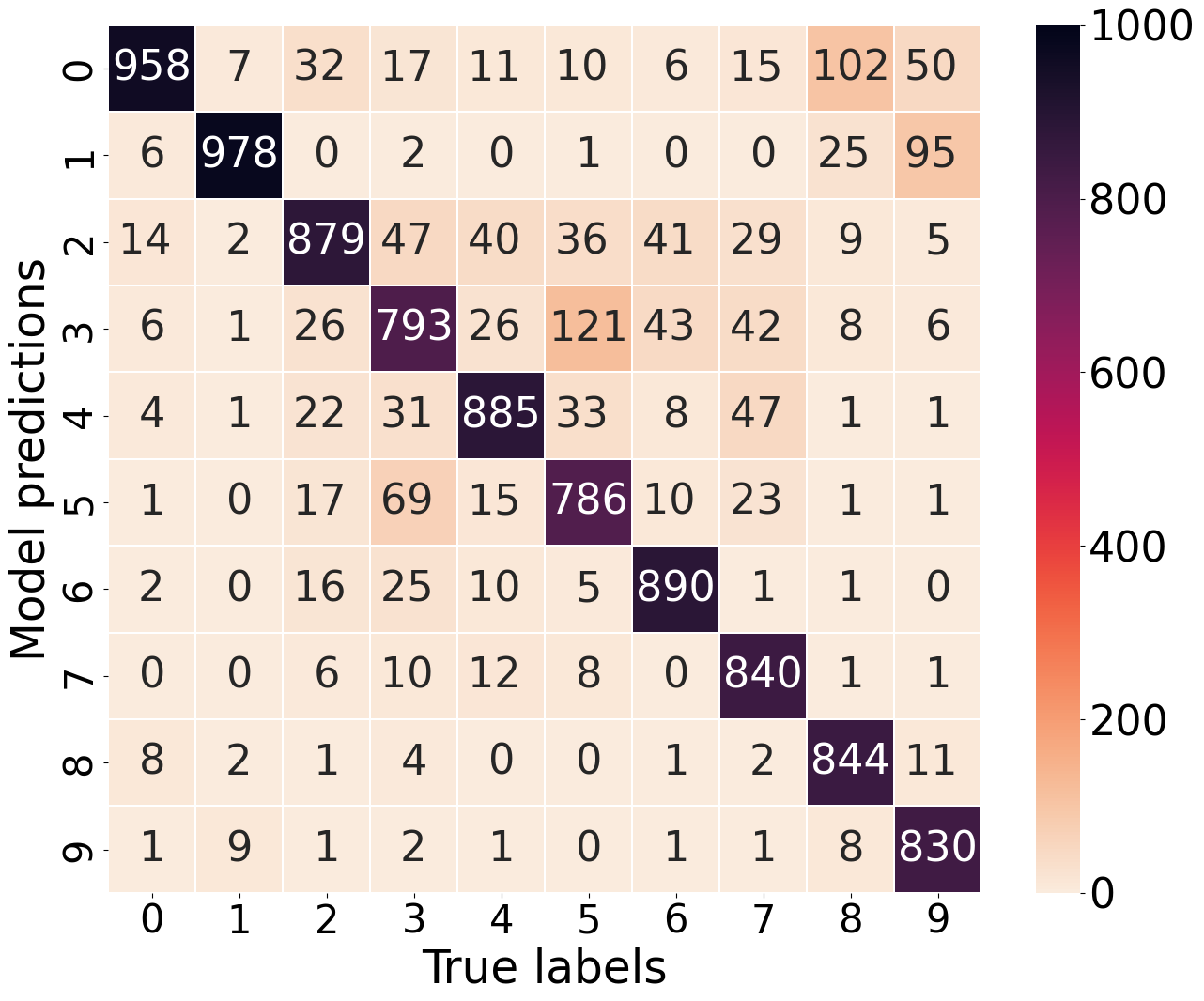}
        \subcaption{ERM, CIFAR10-LT-10 dataset.}
        \label{figure:appendix_confusion_matrix_erm_cifar10_lt_10}
    \end{subfigure}
    \hfill
    \begin{subfigure}[b]{0.32\textwidth}
        \centering
        \includegraphics[width=1.0\textwidth]{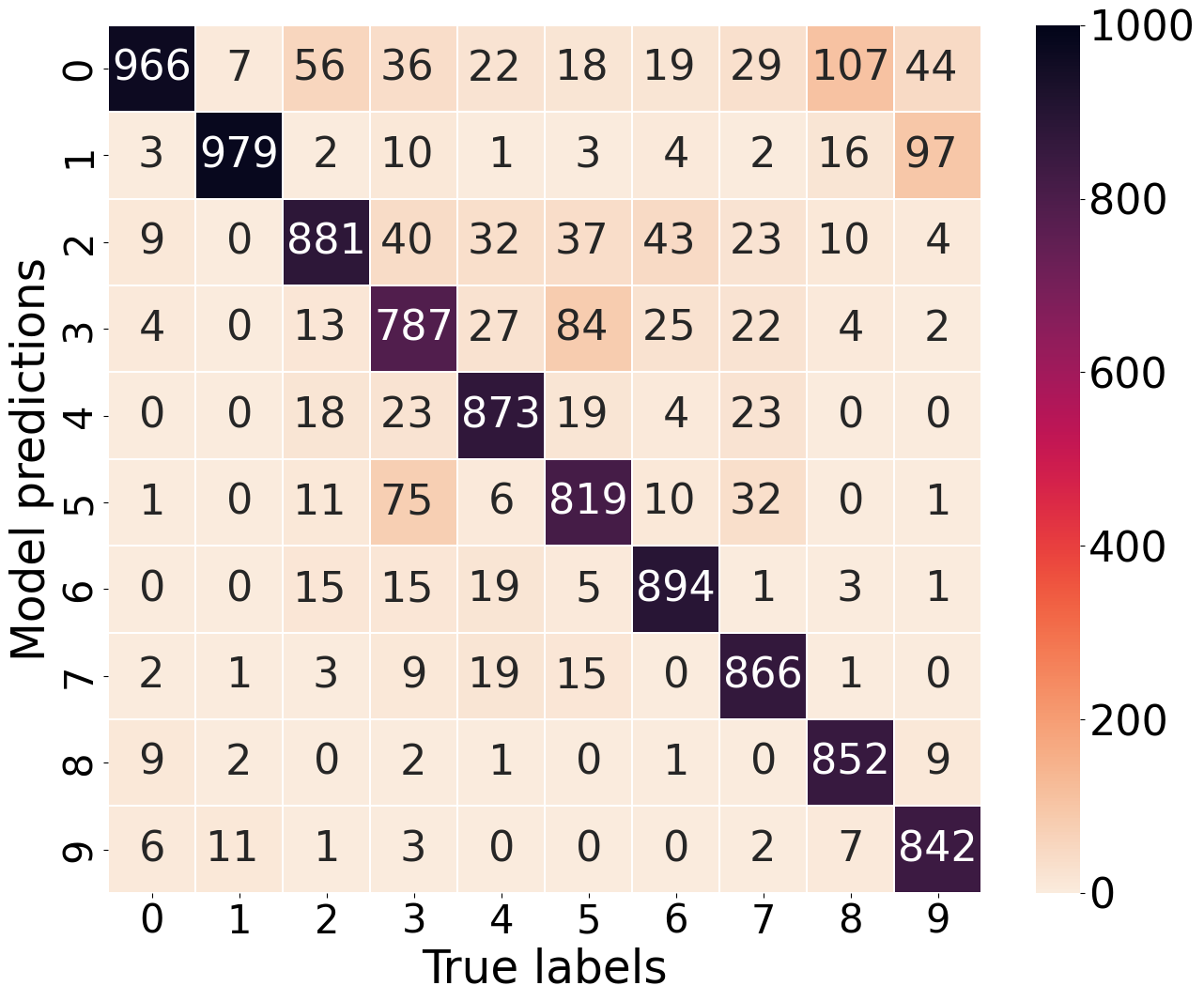}
        \subcaption{Mixup, CIFAR10-LT-10 dataset.}
        \label{figure:appendix_confusion_matrix_mixup_cifar10_lt_10}
    \end{subfigure}
    \hfill
    \begin{subfigure}[b]{0.32\textwidth}
        \centering
        \includegraphics[width=1.0\textwidth]{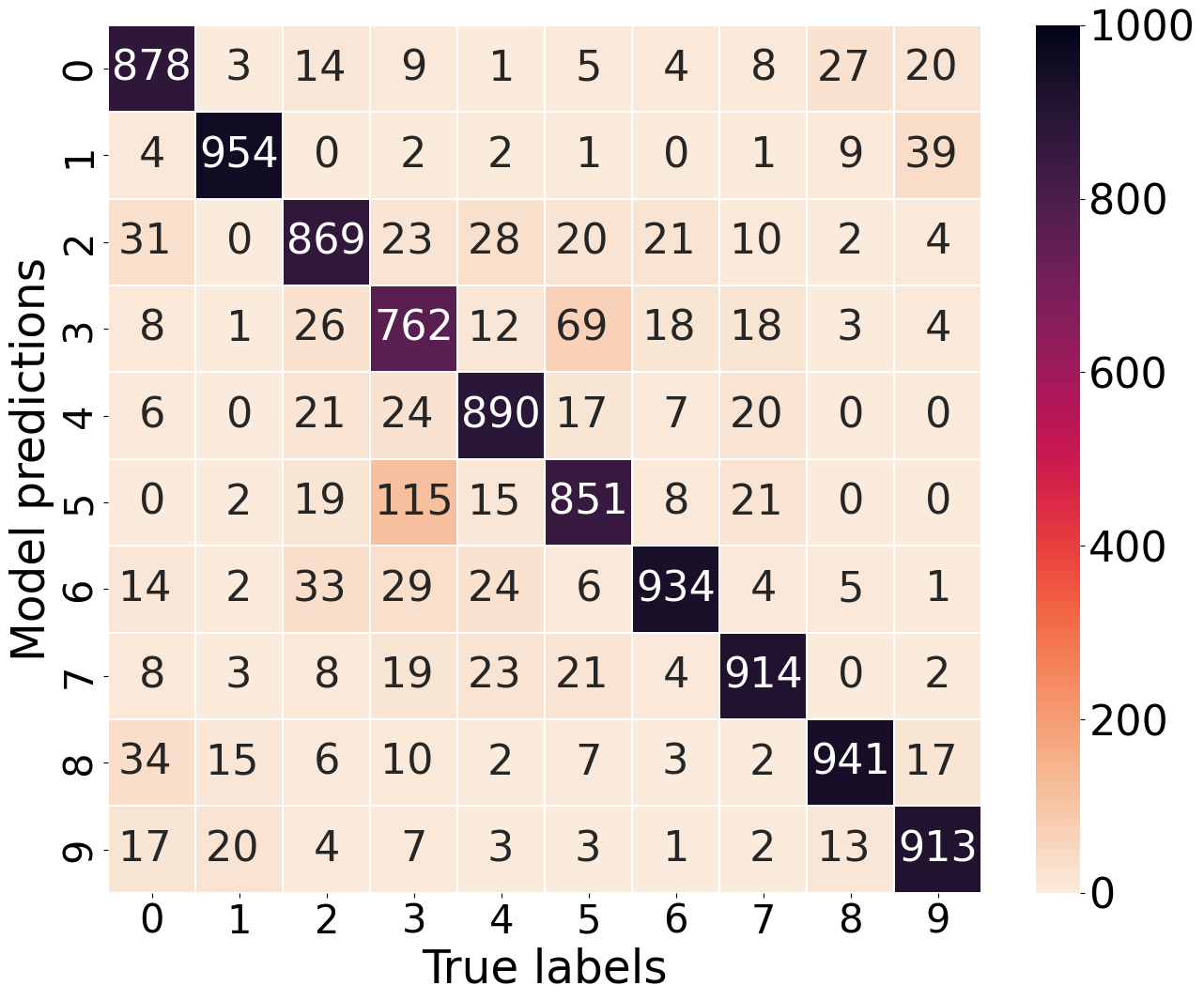}
        \subcaption{CP-Mix, CIFAR10-LT-10 dataset.}
        \label{figure:appendix_confusion_matrix_cpmix_cifar10_lt_10}
    \end{subfigure}
    \hfill
    \caption{
    Confusion matrices of the ERM, Mixup and CP-Mix classifiers trained on CIFAR10-LT datasets.
    }
    \label{figure:appendix_confusion_matrices_cifar10}
\end{figure*}

Figure \ref{figure:appendix_confusion_matrices_cifar100} and Figure \ref{figure:appendix_confusion_matrices_cifar10} demonstrate the confusion matrices of ERM, Mixup and CP-Mix classifiers trained on different CIFAR-LT datasets.
CP-Mix successfully resolves the confusion-pairs and achieves significant performance improvement.

\end{document}